\pdfoutput=1
\documentclass[11pt]{article}

\usepackage{acl}

\usepackage{times}
\usepackage{latexsym}

\usepackage[T1]{fontenc}
\usepackage[utf8]{inputenc}

\usepackage{microtype}

\usepackage{inconsolata}
\usepackage{graphicx}
\usepackage[colorinlistoftodos]{todonotes}
\usepackage{multirow}
\usepackage{colortbl,hhline} 
\usepackage{enumitem}
\setlist[itemize]{noitemsep, nolistsep}
\title{Cross-lingual Editing in Multilingual Language Models}

\author{Himanshu Beniwal$^\dag$$^*$, Kowsik Nandagopan D$^*$, Mayank Singh\\ 
Department of Computer Science and Engineering \\
  Indian Institute of Technology Gandhinagar \\
  \texttt{\{himanshubeniwal, dkowsik, singh.mayank\}@iitgn.ac.in} \\
  }

\begin{document}
\maketitle
\def\thefootnote{$\dag$}\footnotetext{This work is supported by the Prime Minister Research Fellowship.}\def\thefootnote{\arabic{footnote}}
\def\thefootnote{*}\footnotetext{Equal Contribution.}\def\thefootnote{\arabic{footnote}}

\begin{abstract}
The training of large language models (LLMs) necessitates substantial data and computational resources, and updating outdated LLMs entails significant efforts and resources. While numerous model editing techniques (METs) have emerged to efficiently update model outputs without retraining, their effectiveness in multilingual LLMs, where knowledge is stored in diverse languages, remains an underexplored research area. This research paper introduces the cross-lingual model editing (\textbf{XME}) paradigm, wherein a fact is edited in one language, and the subsequent update propagation is observed across other languages. To investigate the XME paradigm, we conducted experiments using BLOOM, mBERT, and XLM-RoBERTa using the two writing scripts: \textit{Latin} (English, French, and Spanish) and \textit{Indic} (Hindi, Gujarati, and Bengali). The results reveal notable performance limitations of state-of-the-art METs under the XME setting, mainly when the languages involved belong to two distinct script families. These findings highlight the need for further research and development of XME techniques to address these challenges. For more comprehensive information, the dataset used in this research and the associated code are publicly available at the following URL\footnote{\url{https://github.com/lingo-iitgn/XME}}.
\end{abstract}

\section{Introduction}
\label{sec:intro}
\begin{figure*}
\begin{center}
\includegraphics[width=\textwidth]{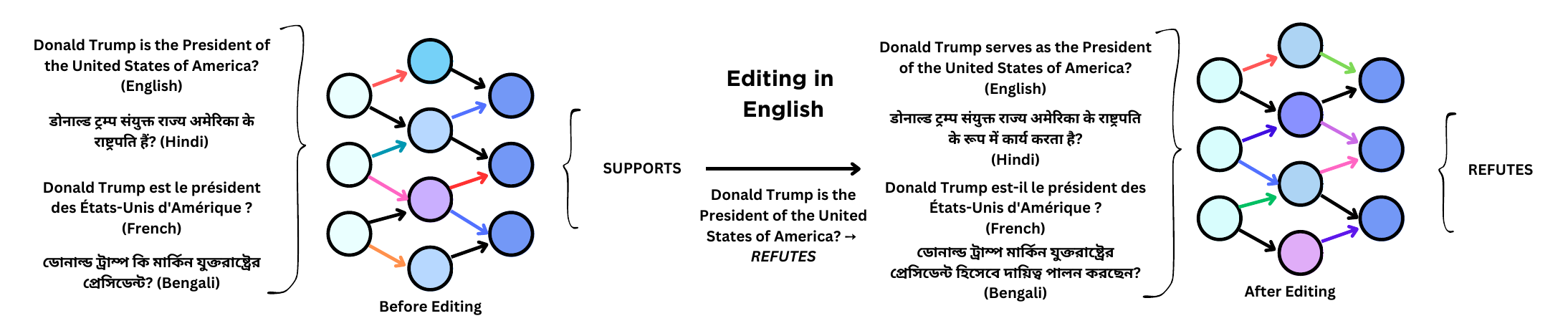}
\end{center}
\caption{XME pipeline: we update a fact in one language (say English) and check whether the same fact is updated in different languages.}
\label{fig:modeledit}
\end{figure*}

The introduction of large language models (LLMs) has revolutionized tasks such as dialogue generation, question-answering, and contextual reasoning \citep{llms, llama, bloom}. LLMs are trained on massive datasets, but this unsupervised data can potentially contain biased or incorrect information. For example, an LLM trained on a dataset of news articles might learn that: \textit{Apple iPhones are the best phones }or that \textit{Mumbai is the capital of India}. This issue becomes problematic because retraining an LLM with equivalent computational power and environmental impact is impractical ~\citep{promptediting, prompting}. To address this problem, researchers have proposed several Model-Editing Techniques (hereafter referred as \textbf{METs}, \citet{knowledgeneurons, knowledgeediting}). METs focus on updating the knowledge within existing LLMs rather than undergoing complete retraining. However, these METs have been evaluated predominantly in monolingual settings, where editing and evaluation occur within a single language, typically English. This paper aims to explore an alternative scenario, as depicted in Figure~\ref{fig:modeledit}. For example, we consider the task of updating a language model (in the English language) to reflect the transition of presidential power from Donald Trump to Joe Biden in the United States, using established model editing techniques. Subsequently, we prompt the updated model with the following French query: \textit{Donald Trump est le présidentdes États-Unis d'Amérique?} (Donald Trump is the President of the United States of America?), expecting the model to correctly predict `\textit{REFUTES}'. We term this new editing paradigm as \textbf{Cross Lingual Model Editing (XME)}.

\iffalse{}
\begin{figure}
\begin{center}
\includegraphics[width=0.9\linewidth]{images/longexample2.pdf}
\end{center}
\caption{An outline for hypernetwork-based model editing technique.}
\label{fig:me_example}
\end{figure}
\fi

We evaluate a specific family of METs that leverage a hypernetwork, an additional model, to update the parameters of a base LLM within the framework of XME. The primary objective is to address the following research questions: \textbf{[Q1]} What is the effectiveness of hypernetwork-based editing techniques in cross-lingual settings? \textbf{[Q2]} Do different architectures store knowledge at different locations? \textbf{[Q3]} How does language selection in the initial fine-tuning stage affect editing performance? \textbf{[Q4]} Is the traditional fine-tuning approach more effective than METs in achieving higher performance in the cross-lingual setting?

In our research, we present the following key contributions:
\begin{itemize}
    \item We explore the cross-lingual editing paradigm on existing METs over two distinct language writing scripts encompassing six languages (both high and low resources).
    \item We uncover a substantial editing performance disparity between monolingual and cross-lingual contexts with exhaustive 9,936 experiments in 69 configurations (Language Pairs x Models x METs).
    \item We provide robust evidence of distinct knowledge localizations in multilingual encoder-only and decoder-only LLMs. 
\end{itemize}

\section{Related Work}
\label{sec:relwork}
We classify previous works into two distinct categories: \textbf{\textit{(i)}} \textit{Parameter-Updating} techniques involve actively updating and modifying the parameters of the LLM. These approaches aim to adapt and fine-tune the LLM's parameters according to the specific requirements of the editing task. These techniques involve the use of additional feed-forward network architectures. Notably, \textbf{KnowledgeEditor} \citep{knowledgeediting} and \textbf{KnowledgeNeurons} \citep{knowledgeneurons} leverage the gradients of the base model and a hypernetwork to identify the weights that require updating \citep{hypernetworks}. Another prominent technique, \textbf{MEND} \citep{mend}, employs gradient updates from multiple feed-forward networks to update the parameters of the base model. Numerous \textit{Locate-then-Edit} techniques, exemplified as \textbf{ROME} \cite{ROME} and \textbf{MEMIT} \citep{memit}, initially localize the knowledge within the model and then update the base model accordingly. 

On the other hand, \textbf{\textit{(ii)}} \textit{Parameter-Preserving} techniques refer to methods that aim to maintain the original parameters of the LLM during the editing process \citep{promptediting, calinet, Patcher}. The focus is on preserving the existing knowledge and capabilities of the LLM while incorporating specific modifications for the desired task. \textbf{SERAC} \citep{serac} incorporates an explicit memory to store edits, enabling the model to reason over them and modulate the predictions of the base model accordingly. Another approach, \textbf{GRACE} by \citet{grace}, introduces a key-value model editor that learns to cache and retrieve activations for selected layers based solely on observed errors during deployment. 
% Additionally, \textbf{Transformer-Patching} \citep{patches} presents a method where a patch, consisting of a set of neurons, is attached to the last Feed-Forward Networks in the transformer block. This patch is responsible for learning the new information while leaving the remaining parameters of the model unchanged. 
% By leveraging heuristics, explicit memory, caching mechanisms, and selective neuron patches, these techniques preserve existing knowledge while accommodating specific updates or edits required for the desired task.

% hypernetwork-based editing techniques have gained significant popularity among parameter updating techniques. 
The preference for hypernetwork-based approaches over other METs arises regarding the effective generability and localization of knowledge, albeit requiring additional memory \citep{Opportunities, crosslanguage}. A study conducted by \citet{does} reveals that localization techniques do not provide further insights into determining the most suitable MLP layer within the base model for overriding an existing stored fact with a new one. Further, the time required to perform an edit in hypernetwork-based techniques is lesser, and the inaccessibility of ROME and MEMIT over different architectures reasons to choose hypernetwork-based techniques over other METs in our experiments \citep{Opportunities}.

\section{Cross-lingual Model Editing (XME)}
\label{sec:prob}
 The cross-lingual model editing problem can be explained by leveraging notations from monolingual model editing. Given a fine-tuned model $f$ with its parameter $\theta$, the prediction or label $y$ can be computed as $y = f(x; \theta)$, where $x$ represents the input sentence. Our objective is to update the model's parameter to $\theta'$ in order to modify the label for input $x$ to a new value $a$, denoted as $a = f(x; \theta')$. However, for the remaining information $\hat{x}$ where $\hat{x} \neq x$, the label remains unchanged as $y$. Let's consider an example: when presented with input $x$ as \textit{Donald Trump is the President of the USA?} and its semantically equivalent input $x'$ as \textit{USA's President is Donald Trump} a fact verification model $f$ outputs $y$ as ``SUPPORTS''. Now, assuming that the fact is updated and model parameters are changed from $\theta$ to $\theta'$, for the same inputs $x$ and $x'$, the updated output becomes `REFUTES' ($a$), where $a = f(x, \theta') = f(x', \theta')$. Furthermore, the unrelated information $\hat{x}$ remains the same as before editing; for instance, \textit{The capital of France is Paris} should still yield the answer ``SUPPORTS''. Therefore, $y = f(\hat{x}; \theta) = f(\hat{x}; \theta')$. In contrast to the monolingual model editing, in XME, the inputs $x$, $x'$, and $\hat{x}$ belong to different languages. 

\iffalse{}
\begin{table}[]
\begin{tabular}{ccccc}
\hline
\textbf{\textbf{Lang}} & \textbf{\textbf{AL$_b$}} & \textbf{\textbf{AL$_m$}} & \textbf{\textbf{AL$_x$}} & \textbf{\textbf{Train set}} \\ \hline
en            & 11.25 & 10.67 & 11.87 & 104966 \\
hi            & 14.4  & 18.04 & 15.69 & 103191 \\
es            & 12.25 & 12.53 & 14.07 & 104965 \\
fr            & 10.5  & 10.6  & 12.79 & 104966 \\
bn            & 13.58 & 20.72 & 17.61 & 104966 \\
gu            & 15.93 & 23.86 & 18.07 & 104966 \\
mix           & 11.25 & 10.67 & 11.25 & 102922 \\
$Inv_{b}$ & 11.25 & -     & -     & 104504 \\
$Inv_{x}$   & -     & -     & 11.95 & 104966 \\ \hline
\end{tabular}
\caption{\label{detailsdataset}
Dataset statistics in different languages. The average length (AL) is represented as AL$_b$, AL$_m$, and AL$_x$ for the BLOOM, mBERT, and XLM-RoBERTa models. The inverse proportion used for BLOOM and XLM-RoBERTa is denoted using $Inv_{b}$ and $Inv_{x}$. Lastly, the size of validation and test remains 10444 and 1193 in all the languages. 
}
\end{table}
\fi

\section{Experiments}
\label{sec:exp}
This section details the experiments performed for XME and highlights the dataset, architectures, and evaluation strategies.

\subsection{Language Selection}
\label{sec:lang_sel}
We have selected a diverse set of languages from the two distinct scripts: \textit{Latin} and \textit{Indic}. From the \textit{Latin} branch family, we have chosen three widely spoken languages: English (\textbf{en}), French (\textbf{fr}), and Spanish (\textbf{es}). These languages have significant global influence and are among the top 10 most spoken languages worldwide \citep{lobachev2008top}. Additionally, we have included three languages from the \textit{Indic} script family: Hindi (\textbf{hi}), Bengali (\textbf{bn}), and Gujarati (\textbf{gu}). Hindi and Bengali are among the top 10 most widely spoken languages globally.

\subsection{Dataset}
\label{sec:data}
In our experimental setup, we focus on a closed-book fact verification task using a modified version of the binary FEVER dataset \citep{fever}. This modified dataset, as described in \citep{knowledgeediting,mend}, includes the original instances and 1 to 25 human-created semantically similar paraphrases for each instance. The dataset consists of 104,966 training instances and 10,444 validation instances. The facts are updated by flipping the label. There are  1,200 instances with flipped labels that were used for editing and subsequent evaluation. On average, each instance has ten semantically similar paraphrases (refer to \S\ref{Appendix:dataset} for more details). We translate\footnote{The translation was performed using Google's Translate API: \url{https://cloud.google.com/translate}} each training, validation, edited instance and the corresponding paraphrases (originally in \textbf{en}) into five languages described above, creating six snapshots of the same data one for each language. Note that, in our experiments, we performed editing and evaluation on 1193 (out of originally 1200 instances) instances, as the rest led to translation errors.
\par \noindent \textbf{Quality Assessment of Translations}: For the five selected languages (other than English), two annotators per language were chosen to verify and annotate the randomly chosen 150 correct translations. All the annotators were native in their assigned languages and fluent in English. The average accuracy and Inter-Annotator Agreement (IAA) over all languages are 88.07\% and 77.8\%, respectively. The details for the average annotator's accuracy and IAA per language are added in \S\ref{sec:annotation}.

\subsection{Pretrained Language Models (PLMs)}
\label{sec:PLMs}
Our research paper investigates the performance of two distinct families of multilingual PLMs: encoder- and decoder-only models. As a representative decoder-only PLM, we choose BLOOM \citep{bloom}. BLOOM is a massive language model trained on the extensive ROOTS corpus \citep{roots}, encompassing 46 diverse natural languages. For the encoder-only category, we selected mBERT (bert-base-multilingual-uncased) \citep{bert} and XLM-RoBERTa \citep{xlmroberta} as representative models based on their well-established performance in multilingual NLP tasks. mBERT, pre-trained on the 104 languages with the largest Wikipedia, offers comprehensive language coverage. On the other hand, XLM-RoBERTa was trained on filtered CommonCrawl data \citep{CommonCrawl}, enabling robust performance across one hundred languages. Considering the limitations imposed by computational resources, we opted to employ a downsized variant of BLOOM, namely BLOOM-560M (hereafter referred to as BLOOM), for our research. Additionally, we utilized uncased versions of mBERT and the base-sized model variant of XLM-RoBERTa in our experiments. 
% \footnote{To the best of our knowledge, we could not find any other decoder-only multilingual PLM.}
\subsection{Model Editing Techniques (METs)}
\label{sec:METs}
We conducted the experiments on two state-of-the-art hypernetwork-based MET techniques along with the standard fine-tuning technique. The hypernetwork-based MET includes Model Editor Networks using Gradient Decomposition (\textbf{MEND},~\citet{mend})  and Knowledge Editor (\textbf{KE},~\citet{knowledgeediting}). Both techniques used an additional model, referred to as hypernetwork, to update the weights of the base PLM model. The hypernetwork is trained with constrained optimization to modify a fact without affecting the rest of the knowledge. In addition, we employed a standard fine-tuning approach (\textbf{FT}) as a baseline approach, which does not require an additional network for the base PLM update.   

\subsection{Evaluation}
\label{sec:eval}
The above three techniques are evaluated using two metrics as described below: 

\noindent \textbf{The Generability Score ($G_S$)} assesses the ability of the MET to predict updated facts on semantically equivalent inputs accurately. To illustrate this, let's consider an example scenario: initially, given an input $x$ such as \textit{The President of the USA is Donald Trump}, the model predicts a label of `SUPPORTS'. Subsequently, the label for $x$ is updated to `REFUTES'. Following the editing of the model parameters, we consider the edit successful if, when presented with semantically equivalent inputs ($x'$) (e.g., \textit{Donald Trump is the President of the USA}), the model correctly outputs `REFUTES'. $G_S$ quantifies the proportion of successfully edited inputs where the model predicts the updated fact label on the corresponding semantically equivalent input. In our experiments, we randomly select one $x'$ among several semantically equivalent inputs of $x$.

\noindent \textbf{The Specificity Score ($S_S$)} evaluates the MET's ability to avoid updating unrelated information. In this context, we define an unrelated input as $\hat{x}$, where $\hat{x}$ is irrelevant to the editing fact $x$. For instance, let's consider the initial input $x$ as \textit{The President of the USA is Donald Trump}, and the model predicts a label as `SUPPORTS'. Subsequently, the label for $x$ is updated to `REFUTES'. Now, if we present an unrelated input $\hat{x}$, such as \textit{The capital of France is Paris}, the model should still predict `SUPPORTS'. $S_S$ measures the proportion of unrelated inputs for which the model correctly maintains the original prediction label for an irrelevant input.

\noindent It is essential to note that in the metric definitions mentioned above, we have considered $x$, $x'$, and $\hat{x}$ within the same language to keep it simple. However, in the actual XME setting, $x$, $x'$, or $\hat{x}$ can belong to multiple languages simultaneously.

\begin{table*}[htpb]
\centering

\resizebox{\linewidth}{!}{%
\begin{tabular}{>{\hspace{0pt}}m{0.020\linewidth}>{\hspace{0pt}}m{0.045\linewidth}|>{\centering\hspace{0pt}}m{0.069\linewidth}>{\centering\hspace{0pt}}m{0.069\linewidth}>{\centering\hspace{0pt}}m{0.069\linewidth}>{\centering\hspace{0pt}}m{0.069\linewidth}>{\centering\hspace{0pt}}m{0.069\linewidth}>{\centering\hspace{0pt}}m{0.069\linewidth}|>{\centering\hspace{0pt}}m{0.058\linewidth}>{\centering\hspace{0pt}}m{0.058\linewidth}>{\centering\hspace{0pt}}m{0.058\linewidth}>{\centering\hspace{0pt}}m{0.069\linewidth}>{\centering\hspace{0pt}}m{0.069\linewidth}>{\centering\arraybackslash\hspace{0pt}}m{0.069\linewidth}} 
\hline
 &  & \multicolumn{6}{>{\centering\hspace{0pt}}m{0.39\linewidth}|}{$G_S$ ($x'$) $\rightarrow$} & \multicolumn{6}{>{\centering\arraybackslash\hspace{0pt}}m{0.39\linewidth}}{$G_S$ ($x'$) $\rightarrow$} \\ 
\hline\multicolumn{1}{>{\centering\hspace{0pt}}m{0.05\linewidth}}{\textbf{Set}} & \textbf{$x$} $\downarrow$ & \textbf{en} & \textbf{fr} & \textbf{es} & \textbf{hi} & \textbf{gu} & \textbf{bn} & \textbf{en} & \textbf{fr} & \textbf{es} & \textbf{hi} & \textbf{gu} & \textbf{bn} \\ \hline\multicolumn{1}{>{\centering\hspace{0pt}}m{0.05\linewidth}}{\textbf{IL}}& en & \cellcolor[HTML]{3E91D9}\textbf{91.79}& \cellcolor[HTML]{4E9ADC}87.51& \cellcolor[HTML]{4D99DB}87.85& \cellcolor[HTML]{B8D6F1}58.93& \cellcolor[HTML]{D0E4F5}52.56& \cellcolor[HTML]{C6DEF3}55.24& \cellcolor[HTML]{4C99DB}\textbf{87.93}& \cellcolor[HTML]{6BAAE1}79.8& \cellcolor[HTML]{67A8E1}80.72& \cellcolor[HTML]{B4D4F0}59.93& \cellcolor[HTML]{DFEDF8}48.37& \cellcolor[HTML]{BBD8F1}58.26 \\ 
& fr & \cellcolor[HTML]{4193D9}90.86& \cellcolor[HTML]{2B86D5}\textbf{96.9}& \cellcolor[HTML]{3B8FD8}92.54& \cellcolor[HTML]{B9D7F1}58.59& \cellcolor[HTML]{D2E5F6}51.89& \cellcolor[HTML]{C4DDF3}55.83& \cellcolor[HTML]{77B2E4}76.36& \cellcolor[HTML]{4E9ADC}\textbf{87.43}& \cellcolor[HTML]{63A6E0}81.81& \cellcolor[HTML]{BBD8F1}58.26& \cellcolor[HTML]{DCEBF8}49.29& \cellcolor[HTML]{C0DBF2}56.92 \\ 
& es & \cellcolor[HTML]{4494DA}90.19& \cellcolor[HTML]{3E91D9}91.79& \cellcolor[HTML]{318AD6}\textbf{95.22}& \cellcolor[HTML]{B8D6F0}59.09& \cellcolor[HTML]{CFE4F5}52.72& \cellcolor[HTML]{C3DDF3}55.99& \cellcolor[HTML]{75B0E3}77.03& \cellcolor[HTML]{67A8E1}80.81& \cellcolor[HTML]{4D9ADC}\textbf{87.68}& \cellcolor[HTML]{B6D5F0}59.51& \cellcolor[HTML]{DFEDF8}48.37& \cellcolor[HTML]{C2DCF3}56.16 \\ 
& hi & \cellcolor[HTML]{BEDAF2}57.25& \cellcolor[HTML]{B9D7F1}58.59& \cellcolor[HTML]{B5D5F0}59.68& \cellcolor[HTML]{2D87D5}\textbf{96.31}& \cellcolor[HTML]{A6CCED}63.7& \cellcolor[HTML]{88BBE7}71.84& \cellcolor[HTML]{D6E7F7}50.88& \cellcolor[HTML]{CFE3F5}52.89& \cellcolor[HTML]{CEE3F5}52.98& \cellcolor[HTML]{9FC8EC}\textbf{65.8}& \cellcolor[HTML]{DEECF8}48.7& \cellcolor[HTML]{BBD8F1}58.26 \\ 
& gu & \cellcolor[HTML]{D0E4F5}52.64& \cellcolor[HTML]{D1E5F6}52.22& \cellcolor[HTML]{CCE2F4}53.65& \cellcolor[HTML]{8DBEE8}70.41& \cellcolor[HTML]{318AD6}\textbf{95.22}& \cellcolor[HTML]{81B7E6}73.68& \cellcolor[HTML]{D8E8F7}50.46& \cellcolor[HTML]{D3E6F6}51.63& \cellcolor[HTML]{D2E5F6}51.97& \cellcolor[HTML]{CEE3F5}53.06& \cellcolor[HTML]{D4E6F6}51.47& \cellcolor[HTML]{BDD9F2}\textbf{57.59} \\ 
& bn & \cellcolor[HTML]{CAE1F4}54.15& \cellcolor[HTML]{CAE1F4}54.06& \cellcolor[HTML]{C6DEF3}55.24& \cellcolor[HTML]{8ABCE8}71.33& \cellcolor[HTML]{9DC7EB}66.14& \cellcolor[HTML]{2C87D5}\textbf{96.65}& \cellcolor[HTML]{D9E9F7}49.96& \cellcolor[HTML]{D3E6F6}51.8& \cellcolor[HTML]{D4E6F6}51.55& \cellcolor[HTML]{CCE2F5}53.56& \cellcolor[HTML]{DDEBF8}49.04& \cellcolor[HTML]{A0C8EC}\textbf{65.55} \\ 
 \hline 
\multicolumn{1}{>{\centering\hspace{0pt}}m{0.05\linewidth}}{\textbf{ML}}& en & \cellcolor[HTML]{2C87D5}\textbf{96.56}& \cellcolor[HTML]{358CD7}94.13& \cellcolor[HTML]{328AD6}94.97& \cellcolor[HTML]{7BB3E5}75.44& \cellcolor[HTML]{A9CEEE}62.95& \cellcolor[HTML]{87BBE7}72.09& \cellcolor[HTML]{398ED8}\textbf{93.04}& \cellcolor[HTML]{4293D9}90.7& \cellcolor[HTML]{4997DB}88.77& \cellcolor[HTML]{A0C8EC}65.55& \cellcolor[HTML]{C7DFF4}54.99& \cellcolor[HTML]{92C0E9}69.32 \\ 
& fr & \cellcolor[HTML]{3E91D9}91.79& \cellcolor[HTML]{2784D4}\textbf{97.99}& \cellcolor[HTML]{2E88D5}96.14& \cellcolor[HTML]{86BAE7}72.34& \cellcolor[HTML]{AACEEE}62.7& \cellcolor[HTML]{90C0E9}69.66& \cellcolor[HTML]{539DDD}86.17& \cellcolor[HTML]{4695DA}\textbf{89.69}& \cellcolor[HTML]{4B98DB}88.27& \cellcolor[HTML]{A4CBED}64.46& \cellcolor[HTML]{C8E0F4}54.57& \cellcolor[HTML]{9AC5EB}66.97 \\ 
& es & \cellcolor[HTML]{4394DA}90.44& \cellcolor[HTML]{338BD6}94.72& \cellcolor[HTML]{2884D4}\textbf{97.65}& \cellcolor[HTML]{86BAE7}72.51& \cellcolor[HTML]{AACFEE}62.61& \cellcolor[HTML]{8EBEE8}70.33& \cellcolor[HTML]{569EDD}85.41& \cellcolor[HTML]{4796DA}\textbf{89.44}& \cellcolor[HTML]{4897DA}89.1& \cellcolor[HTML]{A5CBED}64.21& \cellcolor[HTML]{C7DFF4}54.82& \cellcolor[HTML]{9FC8EC}65.72 \\ 
& hi & \cellcolor[HTML]{B5D4F0}59.85& \cellcolor[HTML]{A8CDED}63.29& \cellcolor[HTML]{A1C9EC}65.21& \cellcolor[HTML]{2B86D5}\textbf{96.9}& \cellcolor[HTML]{529CDC}86.5& \cellcolor[HTML]{4D99DB}87.76& \cellcolor[HTML]{C5DEF3}55.41& \cellcolor[HTML]{B7D6F0}59.35& \cellcolor[HTML]{BBD8F1}58.26& \cellcolor[HTML]{80B6E5}74.1& \cellcolor[HTML]{8EBFE8}70.16& \cellcolor[HTML]{7BB4E5}\textbf{75.27} \\ 
& gu & \cellcolor[HTML]{CCE2F5}53.48& \cellcolor[HTML]{CAE0F4}54.23& \cellcolor[HTML]{C2DCF2}56.41& \cellcolor[HTML]{61A5DF}82.31& \cellcolor[HTML]{2E88D5}\textbf{96.14}& \cellcolor[HTML]{4796DA}89.27& \cellcolor[HTML]{C5DEF3}55.49& \cellcolor[HTML]{BDD9F1}57.75& \cellcolor[HTML]{C0DBF2}56.92& \cellcolor[HTML]{82B7E6}73.6& \cellcolor[HTML]{AACEEE}62.7& \cellcolor[HTML]{76B1E4}\textbf{76.61} \\ 
& bn & \cellcolor[HTML]{C4DDF3}55.66& \cellcolor[HTML]{BDD9F2}57.59& \cellcolor[HTML]{B6D5F0}59.43& \cellcolor[HTML]{61A5DF}82.4& \cellcolor[HTML]{509BDC}86.92& \cellcolor[HTML]{2A86D5}\textbf{97.15}& \cellcolor[HTML]{CBE1F4}53.9& \cellcolor[HTML]{C1DBF2}56.66& \cellcolor[HTML]{C5DEF3}55.57& \cellcolor[HTML]{86BAE7}72.42& \cellcolor[HTML]{83B8E6}\textbf{73.26}& \cellcolor[HTML]{8BBDE8}71.08 \\ 
 \hline 
\multicolumn{1}{>{\centering\hspace{0pt}}m{0.05\linewidth}}{\textbf{LL}}& en & \cellcolor[HTML]{2180D3}\textbf{99.67}& \cellcolor[HTML]{2381D3}99.08& \cellcolor[HTML]{2281D3}99.25& \cellcolor[HTML]{8ABCE8}71.33& \cellcolor[HTML]{B4D4F0}59.93& \cellcolor[HTML]{A5CCED}64.04& \cellcolor[HTML]{549DDD}\textbf{85.83}& \cellcolor[HTML]{6EACE2}78.79& \cellcolor[HTML]{6AAAE1}79.97& \cellcolor[HTML]{BBD8F1}58.09& \cellcolor[HTML]{DFECF8}48.53& \cellcolor[HTML]{A8CDED}63.2 \\ 
& fr & \cellcolor[HTML]{4A98DB}88.43& \cellcolor[HTML]{2080D3}\textbf{99.83}& \cellcolor[HTML]{2482D3}98.91& \cellcolor[HTML]{8FBFE9}69.91& \cellcolor[HTML]{BBD8F1}58.09& \cellcolor[HTML]{A8CDED}63.37& \cellcolor[HTML]{9EC8EB}65.97& \cellcolor[HTML]{4896DA}\textbf{89.19}& \cellcolor[HTML]{71AEE2}78.21& \cellcolor[HTML]{B7D6F0}59.26& \cellcolor[HTML]{DEECF8}48.7& \cellcolor[HTML]{A4CBED}64.46 \\ 
& es & \cellcolor[HTML]{79B2E4}75.94& \cellcolor[HTML]{4293D9}90.78& \cellcolor[HTML]{338BD6}\textbf{94.64}& \cellcolor[HTML]{AACEEE}62.87& \cellcolor[HTML]{BFDAF2}57.17& \cellcolor[HTML]{B7D6F0}59.18& \cellcolor[HTML]{A4CBED}64.46& \cellcolor[HTML]{7DB5E5}74.94& \cellcolor[HTML]{4F9ADC}\textbf{87.26}& \cellcolor[HTML]{B1D2EF}60.86& \cellcolor[HTML]{DDEBF8}49.04& \cellcolor[HTML]{9CC6EB}66.55 \\ 
& hi & \cellcolor[HTML]{B7D6F0}59.26& \cellcolor[HTML]{7AB3E4}75.78& \cellcolor[HTML]{72AEE3}77.87& \cellcolor[HTML]{2080D3}\textbf{100.0}& \cellcolor[HTML]{4394DA}90.36& \cellcolor[HTML]{3F92D9}91.45& \cellcolor[HTML]{CEE3F5}53.06& \cellcolor[HTML]{CCE2F5}53.48& \cellcolor[HTML]{CBE1F4}\textbf{53.9}& \cellcolor[HTML]{F1F7FC}43.59& \cellcolor[HTML]{DFEDF8}48.45& \cellcolor[HTML]{DCEBF8}49.2 \\ 
& gu & \cellcolor[HTML]{CEE3F5}53.06& \cellcolor[HTML]{BAD8F1}58.42& \cellcolor[HTML]{9DC7EB}66.22& \cellcolor[HTML]{559EDD}85.5& \cellcolor[HTML]{2381D3}\textbf{99.16}& \cellcolor[HTML]{4494DA}90.11& \cellcolor[HTML]{D5E7F6}51.21& \cellcolor[HTML]{CEE3F5}\textbf{53.14}& \cellcolor[HTML]{CEE3F5}52.98& \cellcolor[HTML]{D7E8F7}50.71& \cellcolor[HTML]{D8E9F7}50.29& \cellcolor[HTML]{EAF3FA}45.52 \\ 
& bn & \cellcolor[HTML]{C3DCF3}56.08& \cellcolor[HTML]{9FC8EC}65.72& \cellcolor[HTML]{93C2E9}68.82& \cellcolor[HTML]{4394D9}90.53& \cellcolor[HTML]{358CD7}94.22& \cellcolor[HTML]{2180D3}\textbf{99.67}& \cellcolor[HTML]{CFE4F5}52.72& \cellcolor[HTML]{CAE1F4}\textbf{54.15}& \cellcolor[HTML]{CDE2F5}53.4& \cellcolor[HTML]{E8F1FA}46.19& \cellcolor[HTML]{E1EEF9}47.86& \cellcolor[HTML]{E3EFF9}47.53 \\ 
 \hline 
\multicolumn{1}{>{\centering\hspace{0pt}}m{0.05\linewidth}}{\textbf{RL}}& en & \cellcolor[HTML]{3E91D9}\textbf{91.79}& \cellcolor[HTML]{5BA1DE}84.07& \cellcolor[HTML]{509BDC}86.84& \cellcolor[HTML]{A1C9EC}65.13& \cellcolor[HTML]{C4DDF3}55.74& \cellcolor[HTML]{A7CDED}63.54& \cellcolor[HTML]{4997DB}\textbf{88.94}& \cellcolor[HTML]{549DDD}85.83& \cellcolor[HTML]{549EDD}85.75& \cellcolor[HTML]{C9E0F4}54.32& \cellcolor[HTML]{D5E7F6}51.05& \cellcolor[HTML]{A9CEEE}62.95 \\ 
& fr & \cellcolor[HTML]{519CDC}86.76& \cellcolor[HTML]{398ED7}\textbf{93.21}& \cellcolor[HTML]{509BDC}86.92& \cellcolor[HTML]{B8D6F1}59.01& \cellcolor[HTML]{CCE2F5}53.56& \cellcolor[HTML]{BDD9F2}57.5& \cellcolor[HTML]{61A5DF}82.31& \cellcolor[HTML]{4B98DB}\textbf{88.35}& \cellcolor[HTML]{579FDD}85.16& \cellcolor[HTML]{CDE2F5}53.4& \cellcolor[HTML]{D0E4F5}52.64& \cellcolor[HTML]{AFD1EF}61.44 \\ 
& es & \cellcolor[HTML]{529CDD}86.34& \cellcolor[HTML]{5EA3DF}83.24& \cellcolor[HTML]{3C8FD8}\textbf{92.46}& \cellcolor[HTML]{B6D5F0}59.43& \cellcolor[HTML]{CCE2F5}53.48& \cellcolor[HTML]{C0DBF2}56.83& \cellcolor[HTML]{66A8E0}80.97& \cellcolor[HTML]{60A4DF}82.73& \cellcolor[HTML]{4D99DB}\textbf{87.85}& \cellcolor[HTML]{CEE3F5}53.06& \cellcolor[HTML]{CCE2F5}53.56& \cellcolor[HTML]{AFD1EF}61.27 \\ 
& hi & \cellcolor[HTML]{B8D7F1}58.84& \cellcolor[HTML]{C3DCF3}56.08& \cellcolor[HTML]{BEDAF2}57.33& \cellcolor[HTML]{3C90D8}\textbf{92.2}& \cellcolor[HTML]{A2CAEC}64.8& \cellcolor[HTML]{94C2EA}68.57& \cellcolor[HTML]{CBE1F4}53.81& \cellcolor[HTML]{C0DBF2}\textbf{56.75}& \cellcolor[HTML]{C1DCF2}56.5& \cellcolor[HTML]{D3E6F6}51.72& \cellcolor[HTML]{CEE3F5}52.98& \cellcolor[HTML]{D2E5F6}51.89 \\ 
& gu & \cellcolor[HTML]{CDE2F5}53.4& \cellcolor[HTML]{D0E4F5}52.56& \cellcolor[HTML]{CDE2F5}53.4& \cellcolor[HTML]{96C3EA}68.15& \cellcolor[HTML]{3C90D8}\textbf{92.2}& \cellcolor[HTML]{88BBE7}71.84& \cellcolor[HTML]{CAE1F4}54.15& \cellcolor[HTML]{C0DBF2}\textbf{56.92}& \cellcolor[HTML]{C2DCF3}56.33& \cellcolor[HTML]{CAE0F4}54.23& \cellcolor[HTML]{FFFFFF}32.86& \cellcolor[HTML]{ECF4FB}45.1 \\ 
& bn & \cellcolor[HTML]{C4DDF3}55.66& \cellcolor[HTML]{CCE2F5}53.56& \cellcolor[HTML]{C7DFF4}54.99& \cellcolor[HTML]{9AC5EB}67.14& \cellcolor[HTML]{9DC7EB}66.3& \cellcolor[HTML]{3A8FD8}\textbf{92.79}& \cellcolor[HTML]{CBE1F4}53.81& \cellcolor[HTML]{C3DCF3}\textbf{56.08}& \cellcolor[HTML]{C3DDF3}55.91& \cellcolor[HTML]{F7FAFD}41.99& \cellcolor[HTML]{E9F2FA}45.77& \cellcolor[HTML]{FFFFFF}37.8 \\ 
 \hline   
\end{tabular}
}
\caption{\label{tab:mend_mBERT_bloom-560M_en_gs}
The table represents $G_S$ for fine-tuned \texttt{mBERT} (left) and \texttt{BLOOM} (right) on `\textbf{en}' dataset using \textbf{MEND}.
}
\end{table*} 

\begin{table*}[htpb]
\centering

\resizebox{\linewidth}{!}{%
\begin{tabular}{>{\hspace{0pt}}m{0.020\linewidth}>{\hspace{0pt}}m{0.045\linewidth}|>{\centering\hspace{0pt}}m{0.069\linewidth}>{\centering\hspace{0pt}}m{0.069\linewidth}>{\centering\hspace{0pt}}m{0.069\linewidth}>{\centering\hspace{0pt}}m{0.069\linewidth}>{\centering\hspace{0pt}}m{0.069\linewidth}>{\centering\hspace{0pt}}m{0.069\linewidth}|>{\centering\hspace{0pt}}m{0.058\linewidth}>{\centering\hspace{0pt}}m{0.058\linewidth}>{\centering\hspace{0pt}}m{0.058\linewidth}>{\centering\hspace{0pt}}m{0.069\linewidth}>{\centering\hspace{0pt}}m{0.069\linewidth}>{\centering\arraybackslash\hspace{0pt}}m{0.069\linewidth}} 
\hline
 &  & \multicolumn{6}{>{\centering\hspace{0pt}}m{0.39\linewidth}|}{$S_S$ ($\hat{x}$) $\rightarrow$} & \multicolumn{6}{>{\centering\arraybackslash\hspace{0pt}}m{0.39\linewidth}}{$S_S$ ($\hat{x}$) $\rightarrow$} \\ 
\hline\multicolumn{1}{>{\centering\hspace{0pt}}m{0.05\linewidth}}{\textbf{Set}} & \textbf{$x$} $\downarrow$ & \textbf{en} & \textbf{fr} & \textbf{es} & \textbf{hi} & \textbf{gu} & \textbf{bn} & \textbf{en} & \textbf{fr} & \textbf{es} & \textbf{hi} & \textbf{gu} & \textbf{bn} \\ \hline\multicolumn{1}{>{\centering\hspace{0pt}}m{0.05\linewidth}}{\textbf{IL}}& en & \cellcolor[HTML]{2C87D5}98.32& \cellcolor[HTML]{2E88D5}98.09& \cellcolor[HTML]{2B86D5}\textbf{98.41}& \cellcolor[HTML]{3089D6}97.76& \cellcolor[HTML]{2D87D5}98.2& \cellcolor[HTML]{328AD6}97.48& \cellcolor[HTML]{A1C9EC}82.52& \cellcolor[HTML]{529CDC}93.23& \cellcolor[HTML]{60A4DF}91.37& \cellcolor[HTML]{2783D4}99.06& \cellcolor[HTML]{2683D4}99.08& \cellcolor[HTML]{2683D4}\textbf{99.1} \\ 
& fr & \cellcolor[HTML]{2985D4}\textbf{98.76}& \cellcolor[HTML]{3089D6}97.72& \cellcolor[HTML]{2B86D5}98.43& \cellcolor[HTML]{2C87D5}98.26& \cellcolor[HTML]{2B86D5}98.45& \cellcolor[HTML]{2F88D6}97.92& \cellcolor[HTML]{82B7E6}86.8& \cellcolor[HTML]{83B8E6}86.61& \cellcolor[HTML]{579FDD}92.52& \cellcolor[HTML]{2281D3}99.62& \cellcolor[HTML]{2281D3}99.64& \cellcolor[HTML]{2281D3}\textbf{99.73} \\ 
& es & \cellcolor[HTML]{2A86D5}\textbf{98.58}& \cellcolor[HTML]{2E88D5}98.07& \cellcolor[HTML]{2D87D5}98.16& \cellcolor[HTML]{2D87D5}98.24& \cellcolor[HTML]{2B86D5}98.51& \cellcolor[HTML]{3089D6}97.76& \cellcolor[HTML]{84B9E6}86.44& \cellcolor[HTML]{4F9BDC}93.57& \cellcolor[HTML]{74AFE3}88.68& \cellcolor[HTML]{2281D3}\textbf{99.67}& \cellcolor[HTML]{2281D3}99.64& \cellcolor[HTML]{2281D3}99.62 \\ 
& hi & \cellcolor[HTML]{2784D4}\textbf{98.99}& \cellcolor[HTML]{2A86D5}98.55& \cellcolor[HTML]{2784D4}98.97& \cellcolor[HTML]{4495DA}95.03& \cellcolor[HTML]{338AD6}97.42& \cellcolor[HTML]{378DD7}96.81& \cellcolor[HTML]{7CB4E5}87.49& \cellcolor[HTML]{398ED8}96.52& \cellcolor[HTML]{4B98DB}94.17& \cellcolor[HTML]{2381D3}99.56& \cellcolor[HTML]{2080D3}\textbf{99.92}& \cellcolor[HTML]{2180D3}99.85 \\ 
& gu & \cellcolor[HTML]{2884D4}98.89& \cellcolor[HTML]{2985D4}98.78& \cellcolor[HTML]{2784D4}\textbf{98.99}& \cellcolor[HTML]{3C90D8}96.17& \cellcolor[HTML]{5FA4DF}91.49& \cellcolor[HTML]{4394DA}95.18& \cellcolor[HTML]{7FB6E5}87.09& \cellcolor[HTML]{3A8FD8}96.4& \cellcolor[HTML]{4B98DB}94.13& \cellcolor[HTML]{2180D3}\textbf{99.85}& \cellcolor[HTML]{91C0E9}84.79& \cellcolor[HTML]{2180D3}99.83 \\ 
& bn & \cellcolor[HTML]{2784D4}98.95& \cellcolor[HTML]{2A85D5}98.62& \cellcolor[HTML]{2784D4}\textbf{99.04}& \cellcolor[HTML]{388DD7}96.71& \cellcolor[HTML]{398ED7}96.63& \cellcolor[HTML]{549DDD}93.0& \cellcolor[HTML]{7BB3E4}87.74& \cellcolor[HTML]{3A8FD8}96.42& \cellcolor[HTML]{4A98DB}94.3& \cellcolor[HTML]{2180D3}\textbf{99.85}& \cellcolor[HTML]{2180D3}99.77& \cellcolor[HTML]{338AD6}97.42 \\ 
 \hline 
\multicolumn{1}{>{\centering\hspace{0pt}}m{0.05\linewidth}}{\textbf{ML}}& en & \cellcolor[HTML]{318AD6}97.61& \cellcolor[HTML]{388ED7}96.69& \cellcolor[HTML]{358CD7}97.13& \cellcolor[HTML]{3189D6}97.65& \cellcolor[HTML]{2E88D5}\textbf{98.01}& \cellcolor[HTML]{358CD7}97.11& \cellcolor[HTML]{E4EFF9}73.55& \cellcolor[HTML]{9AC5EB}83.53& \cellcolor[HTML]{9BC6EB}83.45& \cellcolor[HTML]{378DD7}96.84& \cellcolor[HTML]{368CD7}\textbf{96.94}& \cellcolor[HTML]{368CD7}\textbf{96.94} \\ 
& fr & \cellcolor[HTML]{2F88D5}\textbf{97.97}& \cellcolor[HTML]{3C8FD8}96.23& \cellcolor[HTML]{338BD6}97.38& \cellcolor[HTML]{3089D6}97.84& \cellcolor[HTML]{2F88D6}97.95& \cellcolor[HTML]{368DD7}96.92& \cellcolor[HTML]{A5CCED}82.0& \cellcolor[HTML]{91C0E9}84.74& \cellcolor[HTML]{82B8E6}86.69& \cellcolor[HTML]{2E88D5}97.99& \cellcolor[HTML]{2E88D5}98.01& \cellcolor[HTML]{2E87D5}\textbf{98.11} \\ 
& es & \cellcolor[HTML]{2D87D5}\textbf{98.2}& \cellcolor[HTML]{368CD7}96.94& \cellcolor[HTML]{3A8ED8}96.48& \cellcolor[HTML]{3189D6}97.65& \cellcolor[HTML]{3089D6}97.8& \cellcolor[HTML]{358CD7}97.11& \cellcolor[HTML]{AFD1EF}80.68& \cellcolor[HTML]{83B8E6}86.67& \cellcolor[HTML]{97C4EA}83.93& \cellcolor[HTML]{2A86D5}98.53& \cellcolor[HTML]{2A86D5}\textbf{98.55}& \cellcolor[HTML]{2A86D5}98.53 \\ 
& hi & \cellcolor[HTML]{2884D4}\textbf{98.89}& \cellcolor[HTML]{2B86D5}98.41& \cellcolor[HTML]{2B86D5}98.45& \cellcolor[HTML]{5DA2DF}91.76& \cellcolor[HTML]{64A6E0}90.82& \cellcolor[HTML]{569FDD}92.6& \cellcolor[HTML]{4F9BDC}93.61& \cellcolor[HTML]{3B8FD8}96.33& \cellcolor[HTML]{4696DA}94.78& \cellcolor[HTML]{2583D4}99.25& \cellcolor[HTML]{2281D3}\textbf{99.67}& \cellcolor[HTML]{2583D4}99.22 \\ 
& gu & \cellcolor[HTML]{2784D4}\textbf{99.02}& \cellcolor[HTML]{2985D4}98.66& \cellcolor[HTML]{2985D4}98.74& \cellcolor[HTML]{509BDC}93.46& \cellcolor[HTML]{97C3EA}83.97& \cellcolor[HTML]{60A4DF}91.34& \cellcolor[HTML]{559EDD}92.77& \cellcolor[HTML]{378DD7}96.88& \cellcolor[HTML]{4495DA}95.03& \cellcolor[HTML]{2281D3}\textbf{99.71}& \cellcolor[HTML]{519CDC}93.38& \cellcolor[HTML]{2784D4}98.99 \\ 
& bn & \cellcolor[HTML]{2884D4}\textbf{98.91}& \cellcolor[HTML]{2B86D5}98.41& \cellcolor[HTML]{2B86D5}98.51& \cellcolor[HTML]{4F9ADC}93.67& \cellcolor[HTML]{5EA3DF}91.64& \cellcolor[HTML]{73AFE3}88.77& \cellcolor[HTML]{559EDD}92.77& \cellcolor[HTML]{3B8FD8}96.35& \cellcolor[HTML]{4595DA}94.97& \cellcolor[HTML]{2281D3}\textbf{99.67}& \cellcolor[HTML]{2281D3}99.62& \cellcolor[HTML]{3A8ED8}96.5 \\ 
 \hline 
\multicolumn{1}{>{\centering\hspace{0pt}}m{0.05\linewidth}}{\textbf{LL}}& en & \cellcolor[HTML]{2683D4}\textbf{99.18}& \cellcolor[HTML]{2B86D5}98.39& \cellcolor[HTML]{2C87D5}98.28& \cellcolor[HTML]{2885D4}98.81& \cellcolor[HTML]{2A86D5}98.58& \cellcolor[HTML]{2985D4}98.72& \cellcolor[HTML]{F0F6FC}71.94& \cellcolor[HTML]{67A8E1}90.4& \cellcolor[HTML]{71AEE3}89.0& \cellcolor[HTML]{328AD6}\textbf{97.46}& \cellcolor[HTML]{338AD6}97.4& \cellcolor[HTML]{328AD6}\textbf{97.46} \\ 
& fr & \cellcolor[HTML]{2482D3}\textbf{99.45}& \cellcolor[HTML]{569FDD}92.62& \cellcolor[HTML]{2E88D5}98.01& \cellcolor[HTML]{2C87D5}98.28& \cellcolor[HTML]{2683D4}99.1& \cellcolor[HTML]{2E88D5}98.07& \cellcolor[HTML]{5EA3DF}91.64& \cellcolor[HTML]{549EDD}92.88& \cellcolor[HTML]{4394DA}95.16& \cellcolor[HTML]{2180D3}99.81& \cellcolor[HTML]{2180D3}99.83& \cellcolor[HTML]{2080D3}\textbf{99.87} \\ 
& es & \cellcolor[HTML]{2482D3}\textbf{99.35}& \cellcolor[HTML]{2E87D5}98.11& \cellcolor[HTML]{3D90D8}96.08& \cellcolor[HTML]{2D87D5}98.13& \cellcolor[HTML]{2A85D4}98.64& \cellcolor[HTML]{2F88D5}97.97& \cellcolor[HTML]{5BA1DE}91.97& \cellcolor[HTML]{4394DA}95.2& \cellcolor[HTML]{539DDD}93.08& \cellcolor[HTML]{2281D3}99.73& \cellcolor[HTML]{2180D3}\textbf{99.77}& \cellcolor[HTML]{2180D3}\textbf{99.77} \\ 
& hi & \cellcolor[HTML]{2482D3}\textbf{99.37}& \cellcolor[HTML]{3089D6}97.82& \cellcolor[HTML]{2F88D6}97.88& \cellcolor[HTML]{B7D6F0}79.59& \cellcolor[HTML]{77B1E4}88.27& \cellcolor[HTML]{7FB6E5}87.22& \cellcolor[HTML]{3B8FD8}96.33& \cellcolor[HTML]{368CD7}97.02& \cellcolor[HTML]{3D91D8}95.98& \cellcolor[HTML]{2482D3}99.43& \cellcolor[HTML]{2281D3}99.6& \cellcolor[HTML]{2281D3}\textbf{99.62} \\ 
& gu & \cellcolor[HTML]{2382D3}\textbf{99.52}& \cellcolor[HTML]{2C87D5}98.32& \cellcolor[HTML]{338AD6}97.44& \cellcolor[HTML]{66A8E0}90.51& \cellcolor[HTML]{FFFFFF}69.32& \cellcolor[HTML]{75B0E3}88.54& \cellcolor[HTML]{398ED7}96.63& \cellcolor[HTML]{348BD7}97.23& \cellcolor[HTML]{3C90D8}96.17& \cellcolor[HTML]{2180D3}\textbf{99.77}& \cellcolor[HTML]{4897DB}94.51& \cellcolor[HTML]{2482D3}99.45 \\ 
& bn & \cellcolor[HTML]{2482D3}\textbf{99.33}& \cellcolor[HTML]{2F88D6}97.88& \cellcolor[HTML]{3089D6}97.74& \cellcolor[HTML]{77B1E4}88.27& \cellcolor[HTML]{82B8E6}86.73& \cellcolor[HTML]{F1F7FC}71.86& \cellcolor[HTML]{398ED8}96.58& \cellcolor[HTML]{358CD7}97.11& \cellcolor[HTML]{378DD7}96.81& \cellcolor[HTML]{2180D3}\textbf{99.79}& \cellcolor[HTML]{2784D4}98.99& \cellcolor[HTML]{358BD7}97.17 \\ 
 \hline 
\multicolumn{1}{>{\centering\hspace{0pt}}m{0.05\linewidth}}{\textbf{RL}}& en & \cellcolor[HTML]{3089D6}97.74& \cellcolor[HTML]{368CD7}97.02& \cellcolor[HTML]{338AD6}97.4& \cellcolor[HTML]{328AD6}97.46& \cellcolor[HTML]{2C86D5}\textbf{98.37}& \cellcolor[HTML]{328AD6}97.53& \cellcolor[HTML]{C1DBF2}78.27& \cellcolor[HTML]{78B2E4}88.12& \cellcolor[HTML]{70AEE2}89.12& \cellcolor[HTML]{338BD6}97.36& \cellcolor[HTML]{338AD6}97.4& \cellcolor[HTML]{328AD6}\textbf{97.48} \\ 
& fr & \cellcolor[HTML]{2B86D5}98.43& \cellcolor[HTML]{4092D9}95.62& \cellcolor[HTML]{338BD6}97.32& \cellcolor[HTML]{3089D6}97.76& \cellcolor[HTML]{2A85D4}\textbf{98.64}& \cellcolor[HTML]{328AD6}97.57& \cellcolor[HTML]{92C1E9}84.62& \cellcolor[HTML]{F1F7FC}71.86& \cellcolor[HTML]{C9E0F4}77.26& \cellcolor[HTML]{378DD7}\textbf{96.88}& \cellcolor[HTML]{388ED7}96.67& \cellcolor[HTML]{3E91D9}95.85 \\ 
& es & \cellcolor[HTML]{2C87D5}\textbf{98.34}& \cellcolor[HTML]{328AD6}97.46& \cellcolor[HTML]{388ED7}96.65& \cellcolor[HTML]{3089D6}97.72& \cellcolor[HTML]{2D87D5}98.2& \cellcolor[HTML]{3189D6}97.65& \cellcolor[HTML]{86BAE7}86.21& \cellcolor[HTML]{C4DDF3}77.91& \cellcolor[HTML]{BAD8F1}79.15& \cellcolor[HTML]{3089D6}97.74& \cellcolor[HTML]{3089D6}\textbf{97.8}& \cellcolor[HTML]{328AD6}97.48 \\ 
& hi & \cellcolor[HTML]{2A85D5}\textbf{98.62}& \cellcolor[HTML]{2E88D5}98.01& \cellcolor[HTML]{2D87D5}98.18& \cellcolor[HTML]{4D99DB}93.94& \cellcolor[HTML]{3D90D8}96.0& \cellcolor[HTML]{4695DA}94.87& \cellcolor[HTML]{4D99DB}93.9& \cellcolor[HTML]{549EDD}92.88& \cellcolor[HTML]{549DDD}92.94& \cellcolor[HTML]{2181D3}99.75& \cellcolor[HTML]{2080D3}\textbf{99.92}& \cellcolor[HTML]{2180D3}99.83 \\ 
& gu & \cellcolor[HTML]{2985D4}\textbf{98.76}& \cellcolor[HTML]{2B86D5}98.51& \cellcolor[HTML]{2B86D5}98.45& \cellcolor[HTML]{4393D9}95.28& \cellcolor[HTML]{569EDD}92.71& \cellcolor[HTML]{4A98DB}94.32& \cellcolor[HTML]{4B98DB}94.19& \cellcolor[HTML]{4E9ADC}93.8& \cellcolor[HTML]{4E9ADC}93.71& \cellcolor[HTML]{2080D3}\textbf{99.96}& \cellcolor[HTML]{3B8FD8}96.31& \cellcolor[HTML]{2180D3}99.77 \\ 
& bn & \cellcolor[HTML]{2985D4}\textbf{98.72}& \cellcolor[HTML]{2C87D5}98.32& \cellcolor[HTML]{2E88D5}97.99& \cellcolor[HTML]{4293D9}95.31& \cellcolor[HTML]{3D91D8}95.98& \cellcolor[HTML]{539DDD}93.11& \cellcolor[HTML]{4B99DB}94.09& \cellcolor[HTML]{5AA1DE}92.08& \cellcolor[HTML]{58A0DE}92.44& \cellcolor[HTML]{2080D3}\textbf{99.89}& \cellcolor[HTML]{2080D3}99.87& \cellcolor[HTML]{2C87D5}98.26 \\ 
 \hline  
\end{tabular}
}
\caption{\label{tab:mend_mBERT_bloom-560M_hi}
The table represents $S_S$ for fine-tuned \texttt{mBERT} on the `\textbf{en}' (left) and `\textbf{hi}' (right) dataset using \textbf{MEND}.
}
\end{table*}

\begin{figure*}
\begin{center}
\includegraphics[width=\linewidth]{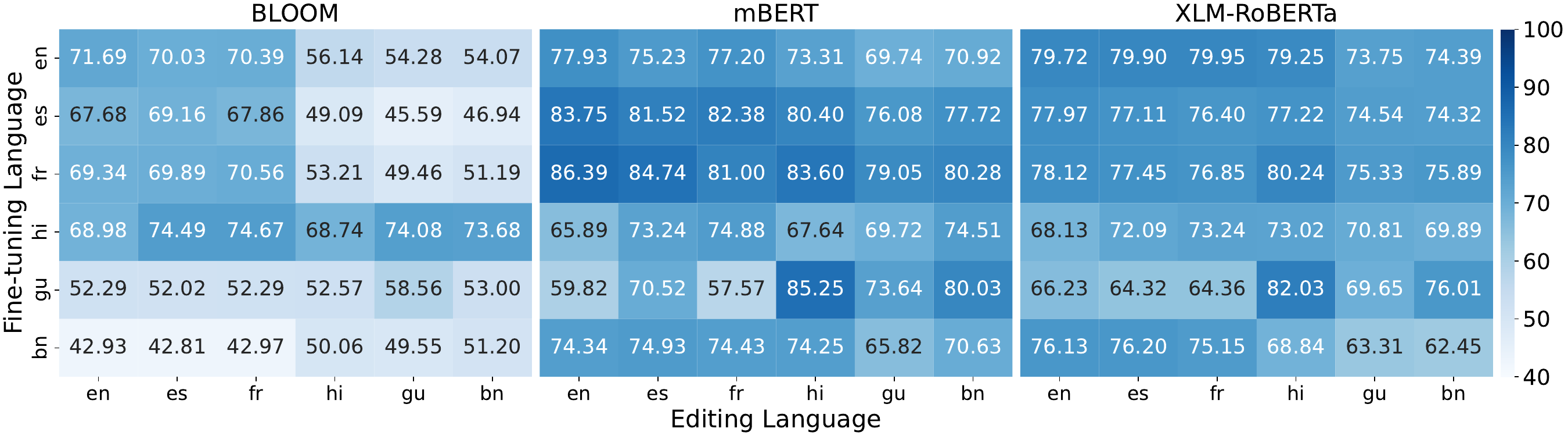}
\end{center}
\caption{The figure illustrates $G_S$ given the editing language (x-axis) and fine-tuning languages (y-axis) for all the three models \texttt{BLOOM} (left), \texttt{mBERT} (middle) and \texttt{XLM-RoBERTa} (right) when edited using \textbf{MEND}.}
\label{fig:mend_wl}
\end{figure*}

\begin{figure*}
\begin{center}
\includegraphics[width=\linewidth]{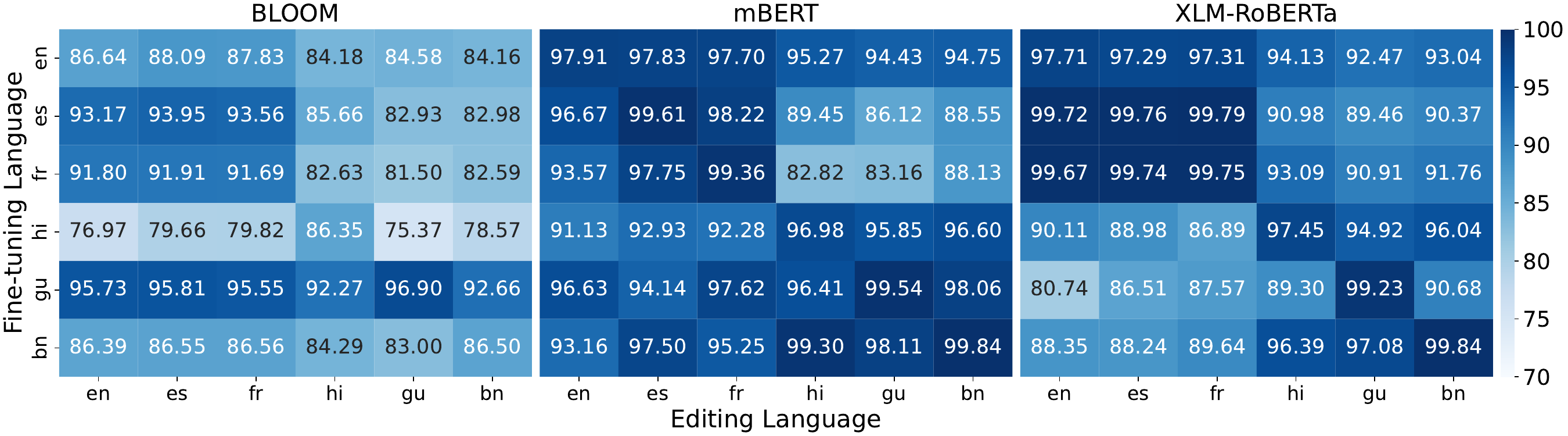}
\end{center}
\caption{The figure illustrates $S_S$ given the editing language (x-axis) and fine-tuning languages (y-axis) for all the three models \texttt{BLOOM} (left), \texttt{mBERT} (middle) and \texttt{XLM-RoBERTa} (right) when edited using \textbf{MEND}.}
\label{fig:mend_wl_loc}
\end{figure*}

\begin{figure*}
\begin{center}
\includegraphics[width=\linewidth]{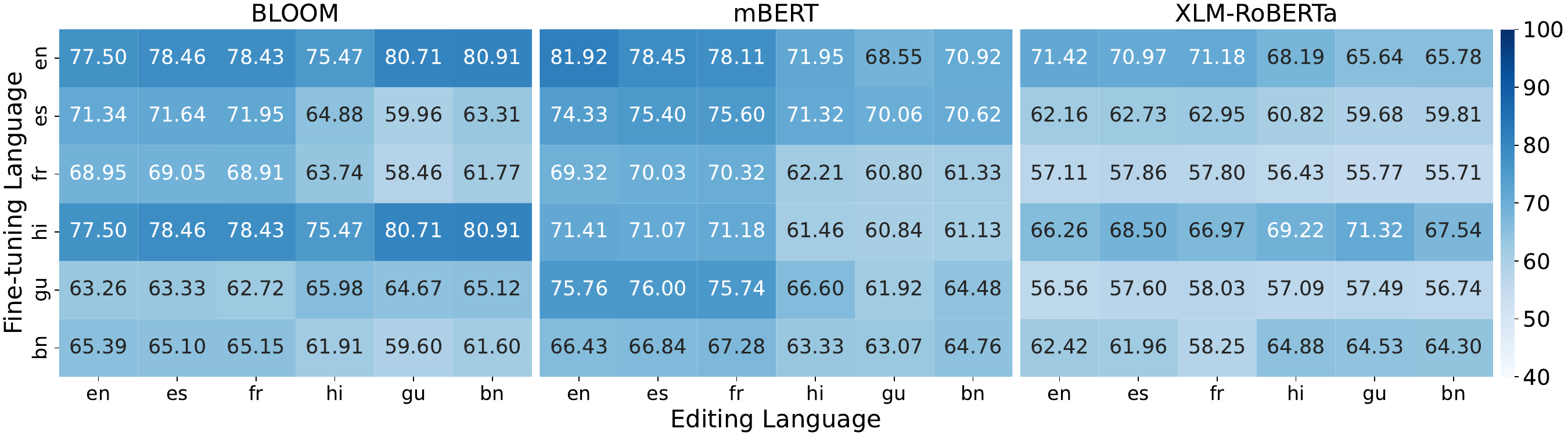}
\end{center}
\caption{The figure illustrates $G_S$ given the editing language (x-axis) and fine-tuning languages (y-axis) for all the three models \texttt{BLOOM} (left), \texttt{mBERT} (middle) and \texttt{XLM-RoBERTa} (right) when edited using \textbf{FT}.}
\label{fig:ft_wl}
\end{figure*}

\begin{figure*}
\begin{center}
\includegraphics[width=\linewidth]{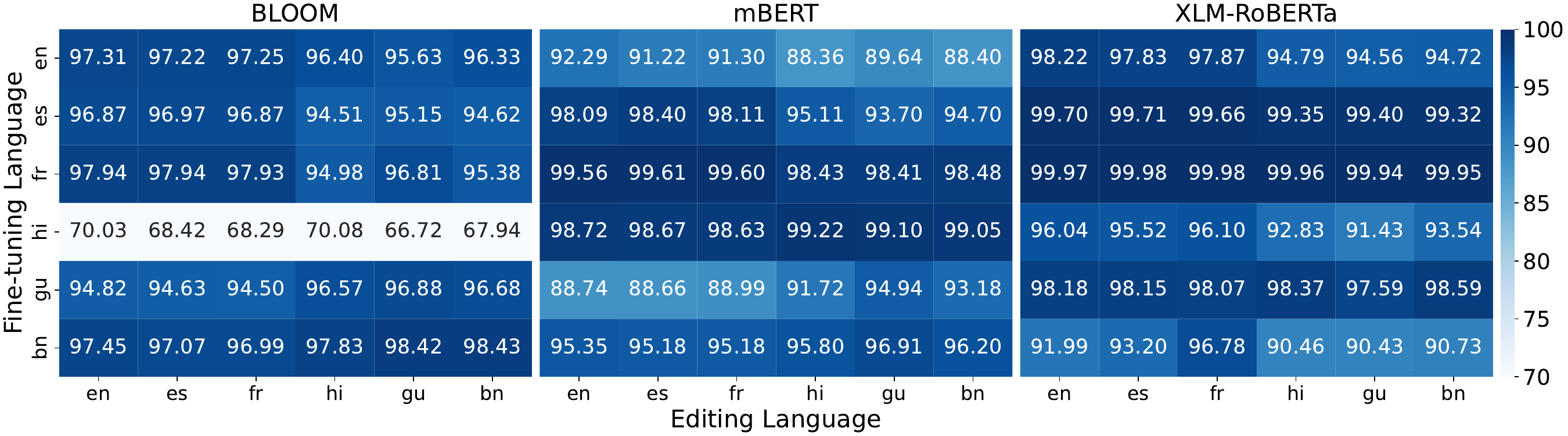}
\end{center}
\caption{The figure illustrates $S_S$ given the editing language (x-axis) and fine-tuning languages (y-axis) for all the three models \texttt{BLOOM} (left), \texttt{mBERT} (middle) and \texttt{XLM-RoBERTa} (right) when edited using \textbf{FT}.}
\label{fig:ft_wl_loc}
\end{figure*}

\subsection{Experimental Settings}
\label{sec:exp_settings}
In our research methodology, we fine-tune the models described in Section~\ref{sec:PLMs} for each specific language. Following the fine-tuning process, we apply model editing techniques, as detailed in Section~\ref{sec:METs}, by passing individual inputs to the fine-tuned models. The performance of these edited models is then evaluated using the metrics defined in Section~\ref{sec:eval}.

To implement the Knowledge Editor and Fine-Tuning techniques, we utilize the implementation provided by \textbf{MEND}~\cite{mend}. Consistent with the experimental settings of \textbf{MEND}, we selectively update only four layers of each PLM. The same set of layers is updated by both \textbf{KE} and \textbf{FT}. For the decoder-only models, we designate layers 1--4 as initial layers (\textbf{IL}), 14--17 as middle layers (\textbf{ML}), 21--24 as last layers (\textbf{LL}), and we randomly select layers 9, 14, 18, and 22 as random layers (\textbf{RL}). Similarly, for the encoder-only models, we assign layers 1-4 as \textbf{IL}, 5--8 as \textbf{ML}, 9--12 as \textbf{LL}, and 3, 5, 7, and 10 as \textbf{RL}. We have utilized the default hyperparameters as implemented in the \textbf{MEND}'s implementation for \textbf{MEND}, \textbf{KE}, and \textbf{FT}. All experiments were completed on 4 V100 GPUs (Each consisting of 32GB).

\section{Results}
\label{sec:results}
In this section, we present and analyze the key findings and address the research questions posed in Introduction Section (see Section~\ref{sec:intro} for more details). To accomplish this, we examine a total of 69\footnote{The combination is (6 languages + mixed configuration + inverse configuration) x 3 models x 3 METs = 72 configurations. Three configurations corresponding to mBERT are unavailable (the inverse proportion for three METs). Hence summing up to 69 configurations.} configurations, which are derived from combining six languages, three PLMs, and three METs. For each configuration, we present the results in tabular form. 
For instance, Table~\ref{tab:mend_mBERT_bloom-560M_en_gs} showcases the performance measured by $G_S$ obtained from fine-tuning the mBERT (left) and BLOOM (right) on an \textbf{en} dataset and subsequently applying the MEND's editing technique. The rows of the table represent the editing languages, while the columns represent the languages used for evaluation. The diagonal values represent monolingual $G_S$, whereas off-diagonal entries show cross-lingual $G_S$. Similarly, Table \ref{tab:mend_mBERT_bloom-560M_hi} showcases the performance measured by $S_S$ for mBERT when fine-tuned on \textbf{en} (left) and \textbf{hi} (right) and edited using MEND. In our experimental analysis, we observe consistent trends for both the MEND and KE techniques. However, due to space limitations, we focus on reporting the results obtained using the MEND approach. The performance scores for the KE technique can be found in \S\ref{sec:ke_all}. Next, we answer the posed research questions. 

\subsection{What is the effectiveness of hypernetwork-based editing techniques in cross-lingual settings?}

Table~\ref{tab:mend_mBERT_bloom-560M_en_gs} and \ref{tab:mend_mBERT_bloom-560M_hi} elucidates notable trends observed in evaluating existing METs. Table~\ref{tab:mend_mBERT_bloom-560M_en_gs} demonstrates high values of $G_S$ (above 90\%) along the diagonal entries, providing empirical evidence for the effectiveness of METs when applied to mBERT in monolingual contexts. Conversely, \textbf{a noticeable decrease in the $G_S$ scores becomes evident as one moves away from the diagonal, indicating the relative inefficiency of METs in cross-lingual scenarios}. Language pairs within the same script family, such as \textbf{en}$\rightarrow$\textbf{es}, \textbf{en}$\rightarrow$\textbf{fr}, or \textbf{hi}$\rightarrow$\textbf{bn}, achieve higher $G_S$ values compared to pairs belonging to different script families, such as \textbf{en}$\rightarrow$\textbf{hi} or \textbf{es}$\rightarrow$\textbf{bn}. The average $G_S$ (excluding the diagonal entries) for editing in the \textit{Latin} family (90.04\%) is significantly higher than in the \textit{Indic} family (78.38\%). However, the two branches do not significantly differ in the average $G_S$ under a monolingual setting. Similar trends are observed for fine-tuning mBERT in other languages (refer to \S\ref{sec:mend_mBERT}, \S\ref{sec:ke_mBERT}, and \S\ref{sec:ft_mBERT} for detailed results). Comparable patterns were also identified for XML-RoBERTa (refer to \S\ref{sec:mend_xlm}, \S\ref{sec:ke_xlm}, \S\ref{sec:ft_xlm} for detailed results). The observations derived from the analysis of the BLOOM model reveal notable distinctions. The metric $G_S$ strongly depends on the fine-tuning language script, irrespective of the employed editing language. Specifically, when examining the \textbf{en} language, a significant disparity in $G_S$ values is observed between the \textit{Latin} and \textit{Indic} script families, as evident in Table~\ref{tab:mend_mBERT_bloom-560M_en_gs}. For instance, the average $G_S$ (including the diagonal entries) for the \textit{Latin} and \textit{Indic} families is 94.14\% and 84.32\%, respectively. Additional results pertaining to BLOOM can be found in \S\ref{sec:mend_bloom}, \S\ref{sec:ke_bloom}, and \S\ref{sec:ft_bloom}.

Unexpectedly, the $S_S$ metric presents contrasting findings compared to the $G_S$ metric. Encoder-only models' $S_S$ mainly depend on the fine-tuning language script irrespective of the editing language. For example, in Table \ref{tab:mend_mBERT_bloom-560M_hi}, average $S_S$ (including the diagonal entries) for the \textit{Latin} family (97.63\%) is sufficiently higher than \textit{Indic} family (91.06\%), when mBERT is finetuned on \textbf{en}. But when fine-tuned on \textbf{hi}, the average $S_S$ for \textit{Indic} family (98.58\%) is higher than the \textit{Latin} family (85.85\%). XLM-RoBERTa follows similar trends (See \S\ref{sec:mend_xlm}, \S\ref{sec:ke_xlm}, \S\ref{sec:ft_xlm} for more details). In contrast, BLOOM shows a very distinct trend. It results in high $S_S$ for the \textit{Latin} script family, irrespective of fine-tuning or editing language selection (refer \S\ref{sec:mend_bloom}, \S\ref{sec:ke_bloom}, and \S\ref{sec:ft_bloom}). \textbf{Lastly, editing and verifying the edit in the same written script family yields better results}. 
\par \noindent \textbf{Inference 1} In our analysis, let us consider that we fine-tune using the ‘en’ dataset, and later we perform the XME. If we look at Table~\ref{tab:mend_mBERT_bloom-560M_en_gs}, for BLOOM (right), the maximum GS for en-en is seen in the Middle layers (93.04\%), while for the last layers, the reported GS is 85.83\%. This shows that it is possible that the model stores the facts at different locations. Similarly, let us consider when we fine-tune using ‘en’ (In the same table) and edit and verify in Spanish (es-es); in this case, the reported GS is 89.1\% in the middle layer and 87.26\% in last layers. The information is significantly (different from nearly 2\%) available across the sets of layers. We have extended the research question by exploring if the fine-tuning language also has any impact on the editing and if it shifts the information from the middle layers to other sets of layers.

\par \noindent \textbf{Inference 2} Referring to Table~\ref{tab:mend_bloom-560M_hi}, we fine-tune the BLOOM model on the ‘hi’ dataset. The GS score for hi-hi in the initial layer (92.37) is higher than the middle layers (85.58\%), which tells us that when we fine-tuned the model on the Hindi dataset, the information is majorly stored in the initial layers rather than our previous assumption of middle layers.

\subsection{Do different architectures store factual knowledge at different locations?}

We have observed that \textbf{different architectures store factual knowledge in distinct locations}. Specifically, in the case of encoder-only models, a significant proportion of factual knowledge is found in the Last Layers (LL). Table~\ref{tab:mend_mBERT_bloom-560M_en_gs} illustrates that the LL exhibits the highest average $G_S$ score (78.74\%) compared to other layer sets (IL=70.23\%, ML=77.93\%, and RL=69.32\%). In contrast, for BLOOM (decoder-based), factual knowledge is concentrated in the Middle Layers (ML). The ML achieves a notably higher average $G_S$ score (69.99\%) than other layer sets (IL=61.16\%, LL=59.19\%, and RL=60.73\%). This finding aligns with the observations made in~\citep{ROME}, which identified similar trends in GPT-2 \citep{gpt2}, decoder-only model \citep{bissaza-cross}. Notably, the initial layers demonstrate the lowest $G_S$ scores for both encoder- and decoder-only models.

\begin{figure}
\begin{center}
\includegraphics[width=0.8\linewidth]{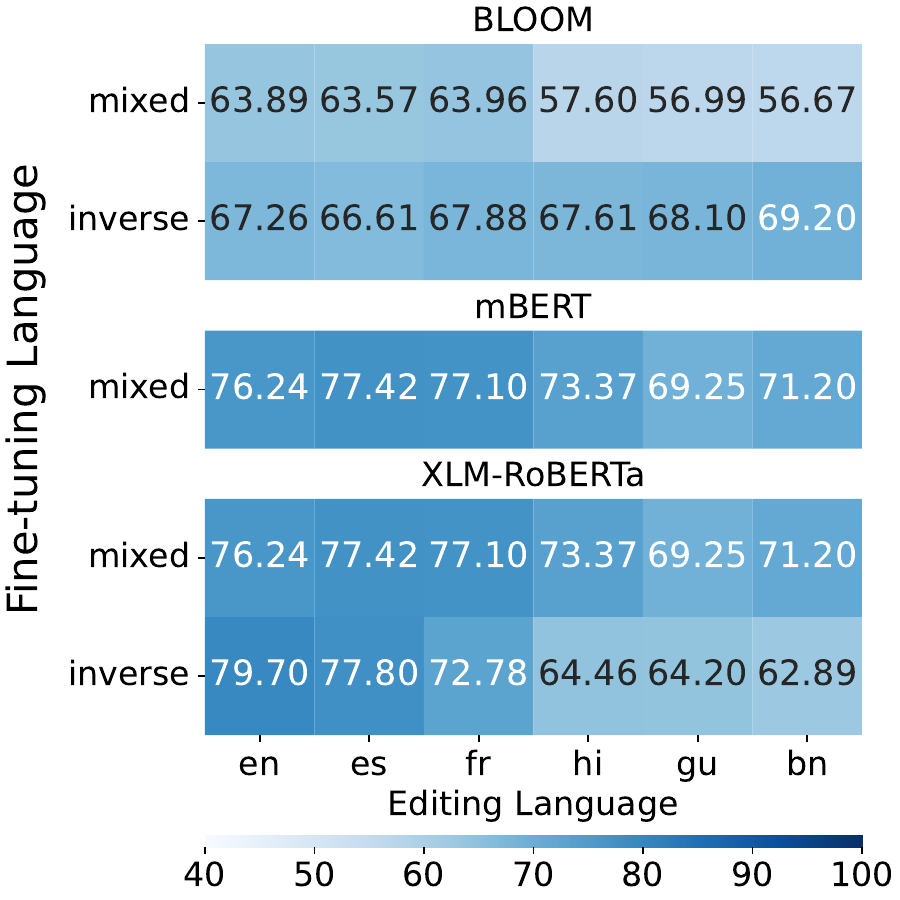}
\end{center}
\caption{The figure illustrates $G_S$ given the editing language (x-axis) and fine-tuning datasets (y-axis) for all the three models \texttt{BLOOM} (top), \texttt{mBERT} (middle) and \texttt{XLM-RoBERTa} (right) when edited using \textbf{MEND}.}
\label{fig:mend_wl_mi}
\end{figure}

\subsection{How does language selection in the initial fine-tuning affect editing performance?}

Figure~\ref{fig:mend_wl} shows the effect of initial fine-tuning performed using six languages. Columns represent average $G_S$ scores for each editing language. As illustrated, language selection during initial fine-tuning significantly impacts the editing performance for the decoder-only model BLOOM. For instance, fine-tuning on the \textit{Latin} script family led to poor $G_S$ for the \textit{Indic} script family. Similar trends can be observed when fine-tuning is performed on \textit{Indic} script families. However, in the latter case, the difference of $G_S$ between the two families is not as high as observed in the former scenario. In the case of encoder-only models, we see a similar performance in both families for \textit{Latin} scripts fine-tuning.\textbf{ In the case of \textit{Indic} family fine-tuning, the performance of \textit{Latin} scripts is marginally poor than that of \textit{Indic} family.} We attribute this to the effect of editing performance on the disproportionate pretraining on different languages.

We performed additional experiments involving two alternative fine-tuning settings. We created two snapshots of the fine-tuning data: (i) \textbf{``mixed''}, which contained an equal distribution of languages, and (ii) \textbf{``inverse''}, where the languages were represented inversely proportional to their respective pretraining language proportions. It is important to note that a single instance of the mixed dataset was generated for PLMs, while the inverse datasets were specific to each PLM. Since BLOOM and XLM-RoBERTa provide language representation information, we only created inverse datasets for these PLMs. Figure~\ref{fig:mend_wl_mi} illustrates the results obtained from the mixed and inverse datasets. Notably, the inverse dataset consistently exhibited performance improvements for the BLOOM model (aka decoder-based). However, the mixed fine-tuning approach performs poorly than the monolingual fine-tuning method. Lastly, in the case of encoder-only models, the mixed and inverse fine-tuning approaches decreased performance compared to the monolingual fine-tuning method.

\begin{figure}
\begin{center}
\includegraphics[width=0.8\linewidth]{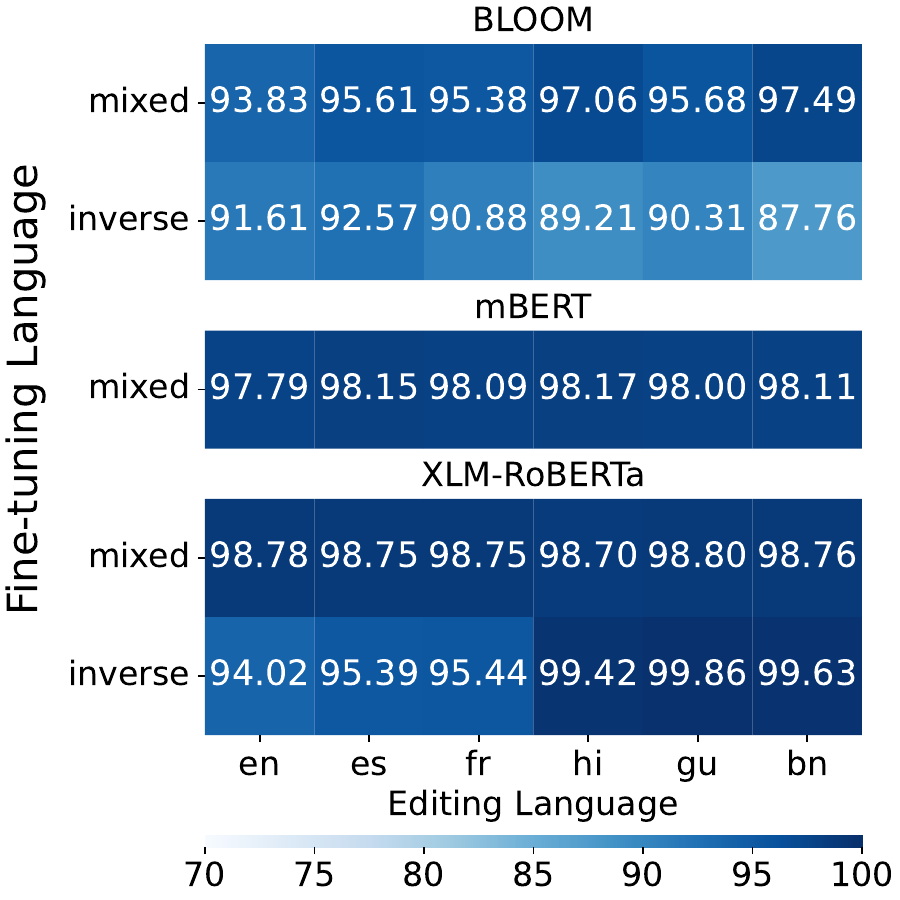}
\end{center}
\caption{The figure illustrates $S_S$ given the editing language (x-axis) and fine-tuning datasets (y-axis) for all the three models \texttt{BLOOM} (top), \texttt{mBERT} (middle) and \texttt{XLM-RoBERTa} (right) when edited using \textbf{MEND}.}
\label{fig:mend_wl_mi_loc}
\end{figure}

Intriguingly, the $S_S$ metric reveals contrasting findings compared to the $G_S$ metric. Figure~\ref{fig:mend_wl_loc} demonstrates that \textbf{the initial fine-tuning significantly impacts the $S_S$ scores of encoder-only models, whereas this observation is not observed for decoder-only models.} Similarly, Figure~\ref{fig:mend_wl_mi_loc} highlights that encoder-only models trained on the mixed dataset exhibit improved $S_S$ scores compared to monolingual fine-tuning. However, the mixed and inverse datasets do not result in any performance gain for the BLOOM model.

\subsection{Is the traditional fine-tuning approach more effective than METs in achieving higher performance in the cross-lingual setting?}

\textbf{Figures ~\ref{fig:ft_wl} and ~\ref{fig:ft_wl_loc} demonstrate that traditional fine-tuning approaches perform comparably to METs in cross-lingual settings}. This observation contrasts the previous claim that shows the significantly low performance of METs in the monolingual setting \citep{crosslanguage, ROME}.

\section{Conclusion and Future Directions} 
\label{sec:con}
Our research focuses on conducting rigorous experiments with state-of-the-art hypernetwork-based model editing techniques within cross-lingual settings. Specifically, we investigate the storage patterns of factual associations in encoder-only and decoder-only models, using two distinct language families as our experimental basis. Additionally, we establish a clear dependency between the fine-tuning language selection and the editing tasks' performance.

To further advance the XME paradigm, we plan to utilize parameter-preserving and localized editing techniques. Furthermore, we intend to extend our investigations to encompass other NLP tasks, such as Machine Translation or question-answering. By expanding our research, we aim to enhance our understanding of the capabilities and limitations of hypernetwork-based model editing techniques in diverse cross-lingual settings.

\section*{Limitations} 
\label{sec:limit}
The performance of METs including \textbf{KN} \citep{knowledgeneurons}, \textbf{SERAC} \citep{serac}, \textbf{CaliNet} \citep{calinet}, \textbf{Transformer-Patching} \citep{patches}, \textbf{KAFT} \citep{KAFT}, \textbf{Patcher} \citep{Patcher}, is limited when the information is distributed across layers. Our experiments' findings indicate that the information in different languages is dispensed across types of architectures. While our work focuses on encoder-based and decoder-based architectures, we intend to incorporate encoder-decoder architectures in future research. The objective is to enhance the localizing and efficient updating of factual information in tasks such as generation, translation, and others. To assess the cross-linguality in METs, we aim to propose a dataset to evaluate whether facts dependent on the edited information also undergo changes. For instance, does the fact `\textit{Where is the President of the USA's hometown?}' also change when we edit the information about the `\textit{President of USA}'. 

% \section*{Potential Risks}
% Individuals or organizations can intentionally delete or alter any information from the LLM's memory. 

\section*{Ethics and Potential Risks}
\label{sec:ethicsec}
The model-editing techniques are designed to edit or delete the information from the LLMs. The editing techniques can be used to modify the model's parameters and can be adversely used. We do not show such harm and intend to show cross-lingual model editing. We carefully adhere to the ethics and guidelines and ensure our work is ethically correct.

\section*{Acknowledgements}
This work is supported by the Prime Minister Research Fellowship (PMRF-1702154) to Himanshu Beniwal. We want to thank Mansi Rana, Anant Kumar, Mihir Patel, Shikhar Nigam, Akbar Ali, Hitesh Lodwal, Indrani Zamindar, Krishna Satish, and Ariana Villegas who helped in verifying the translations. A part of our work was supported by Microsoft's Accelerate Foundation Models Research grant. Lastly, we would like to thank the PARAM Ananta Supercomputing facility under the National Supercomputing Mission coordinated by the Ministry of Electronics and Information Technology (MeitY) and Department of Science and Technology (DST), Government of India, hosted at IIT Gandhinagar.

\bibliography{anthology,custom}

\appendix

\begin{table*}
\centering
\begin{tabular}{llllllllll}
\hline
\textbf{MET}   & \textbf{Model}       & \textbf{en} & \textbf{fr} & \textbf{es} & \textbf{hi} & \textbf{gu} & \textbf{bn} & \textbf{mixed} & \textbf{inverse} \\ \hline
MEND & BLOOM  & \ref{tab:mend_bloom-560M_en}   & \ref{tab:mend_bloom-560M_fr}   &  \ref{tab:mend_bloom-560M_es}  &  \ref{tab:mend_bloom-560M_hi}  & \ref{tab:mend_bloom-560M_gu}   & \ref{tab:mend_bloom-560M_bn}   &   \ref{tab:mend_bloom-560M_mixed}    &    \ref{tab:mend_bloom-560M_inverse}     \\
     & mBERT       & \ref{tab:mend_mBERT_en}   &  \ref{tab:mend_mBERT_fr}  &  \ref{tab:mend_mBERT_es}  &  \ref{tab:mend_mBERT_hi}  &  \ref{tab:mend_mBERT_gu}  & \ref{tab:mend_mBERT_bn}   &  \ref{tab:mend_mBERT_mixed}     &    -     \\
     & XLM-RoBERTa & \ref{tab:mend_XLM-RoBERTa_en}   &   \ref{tab:mend_XLM-RoBERTa_fr}  &  \ref{tab:mend_XLM-RoBERTa_es}   &  \ref{tab:mend_XLM-RoBERTa_hi}   &  \ref{tab:mend_XLM-RoBERTa_gu}   &  \ref{tab:mend_XLM-RoBERTa_bn}   &    \ref{tab:mend_XLM-RoBERTa_mixed}    &   \ref{tab:mend_XLM-RoBERTa_inverse}       \\
KE   & BLOOM  & \ref{tab:ke_bloom-560M_en}   &  \ref{tab:ke_bloom-560M_fr}  &  \ref{tab:ke_bloom-560M_es}  & \ref{tab:ke_bloom-560M_hi}   & \ref{tab:ke_bloom-560M_gu}   & \ref{tab:ke_bloom-560M_bn}   &   \ref{tab:ke_bloom-560M_mixed}    &   \ref{tab:ke_bloom-560M_inverse}      \\
     & mBERT       &  \ref{tab:ke_mBERT_en}  &  \ref{tab:ke_mBERT_fr}  &  \ref{tab:ke_mBERT_es}  & \ref{tab:ke_mBERT_hi}   &  \ref{tab:ke_mBERT_gu}  &  \ref{tab:ke_mBERT_bn}  &   \ref{tab:ke_mBERT_mixed}    &  -       \\
     & XLM-RoBERTa &  \ref{tab:ke_XLM-RoBERTa_en}  &  \ref{tab:ke_XLM-RoBERTa_fr}   &   \ref{tab:ke_XLM-RoBERTa_es}  &   \ref{tab:ke_XLM-RoBERTa_hi}  &   \ref{tab:ke_XLM-RoBERTa_gu}  &   \ref{tab:ke_XLM-RoBERTa_bn}  &    \ref{tab:ke_XLM-RoBERTa_mixed}    &     \ref{tab:ke_XLM-RoBERTa_inverse}     \\
FT   & BLOOM  &  \ref{tab:ft_bloom-560M_en}  &  \ref{tab:ft_bloom-560M_fr}  &  \ref{tab:ft_bloom-560M_es}  &  \ref{tab:ft_bloom-560M_hi}  &  \ref{tab:ft_bloom-560M_gu}  &  \ref{tab:ft_bloom-560M_bn}  &   \ref{tab:ft_bloom-560M_mixed}    &   \ref{tab:ft_bloom-560M_inverse}      \\
     & mBERT       &  \ref{tab:ft_mBERT_en}  &  \ref{tab:ft_mBERT_fr}  &  \ref{tab:ft_mBERT_es}  &  \ref{tab:ft_mBERT_hi}  &  \ref{tab:ft_mBERT_gu}  &  \ref{tab:ft_mBERT_bn}  &  \ref{tab:ft_mBERT_mixed}     &  -      \\
     & XLM-RoBERTa &  \ref{tab:ft_XLM-RoBERTa_en}  &  \ref{tab:ft_XLM-RoBERTa_fr}  &  \ref{tab:ft_XLM-RoBERTa_es}  &  \ref{tab:ft_XLM-RoBERTa_hi}  &  \ref{tab:ft_XLM-RoBERTa_gu}  &  \ref{tab:ft_XLM-RoBERTa_bn}  &   \ref{tab:ft_XLM-RoBERTa_mixed}    &   \ref{tab:ft_XLM-RoBERTa_inverse}     \\ \hline
\end{tabular}
\caption{\label{index_table}
The table contains the index for all the configurations for ME techniques, models, and fine-tuning data.
}
\end{table*}
\section{Appendix}
\label{sec:appendix}

This section contains all the $G_S$ and $S_S$ experiments using different ME techniques for different architectures. 

\begin{figure}
\begin{center}
\includegraphics[width=0.8\linewidth]{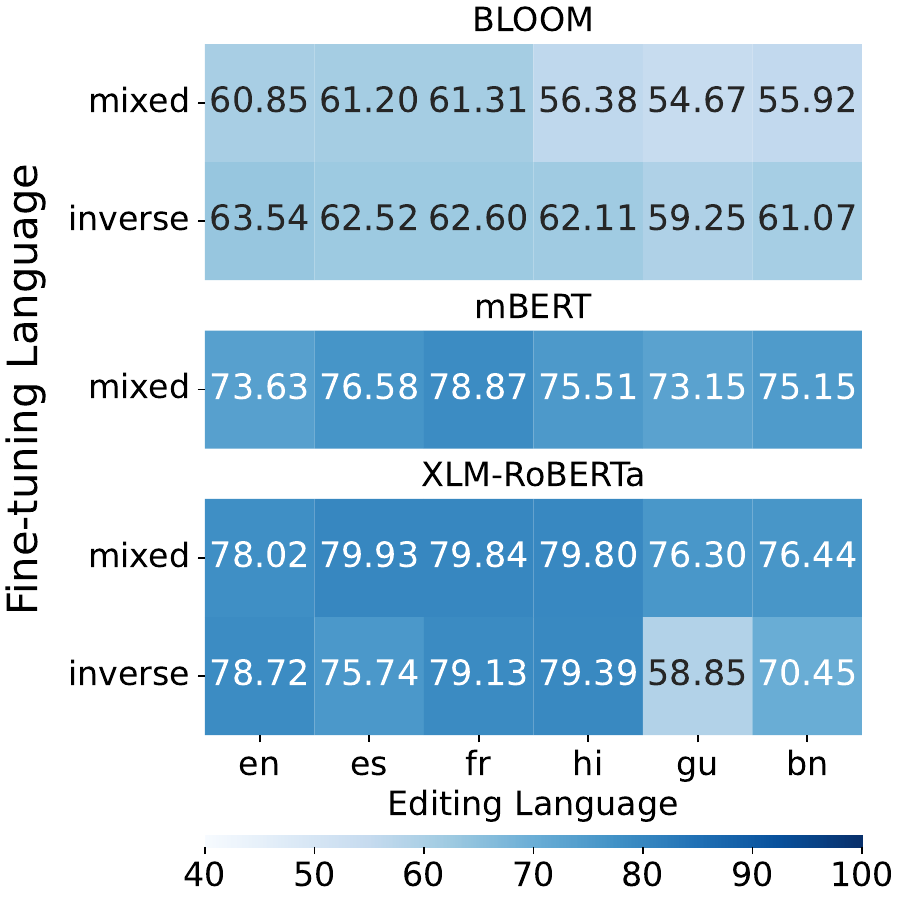}
\end{center}
\caption{The figure illustrates $G_S$ given the editing language (x-axis) and fine-tuning datasets (y-axis) for all the three models \texttt{BLOOM} (top), \texttt{mBERT} (middle) and \texttt{XLM-RoBERTa} (right) when edited using \textbf{KE}.}
\label{fig:ke_wl_mi}
\end{figure}

\begin{figure}
\begin{center}
\includegraphics[width=0.8\linewidth]{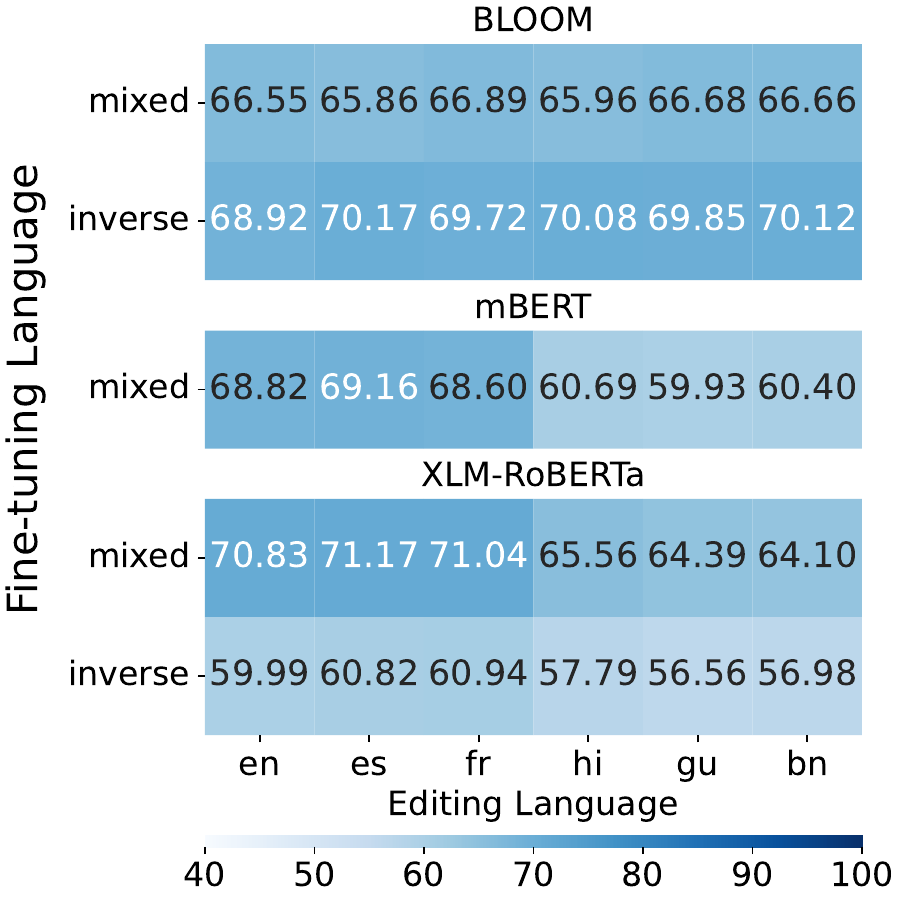}
\end{center}
\caption{The figure illustrates $G_S$ given the editing language (x-axis) and fine-tuning datasets (y-axis) for all the three models \texttt{BLOOM} (top), \texttt{mBERT} (middle) and \texttt{XLM-RoBERTa} (right) when edited using \textbf{FT}.}
\label{fig:ft_wl_mi}
\end{figure}

\begin{figure*}
\begin{center}
\includegraphics[width=\linewidth]{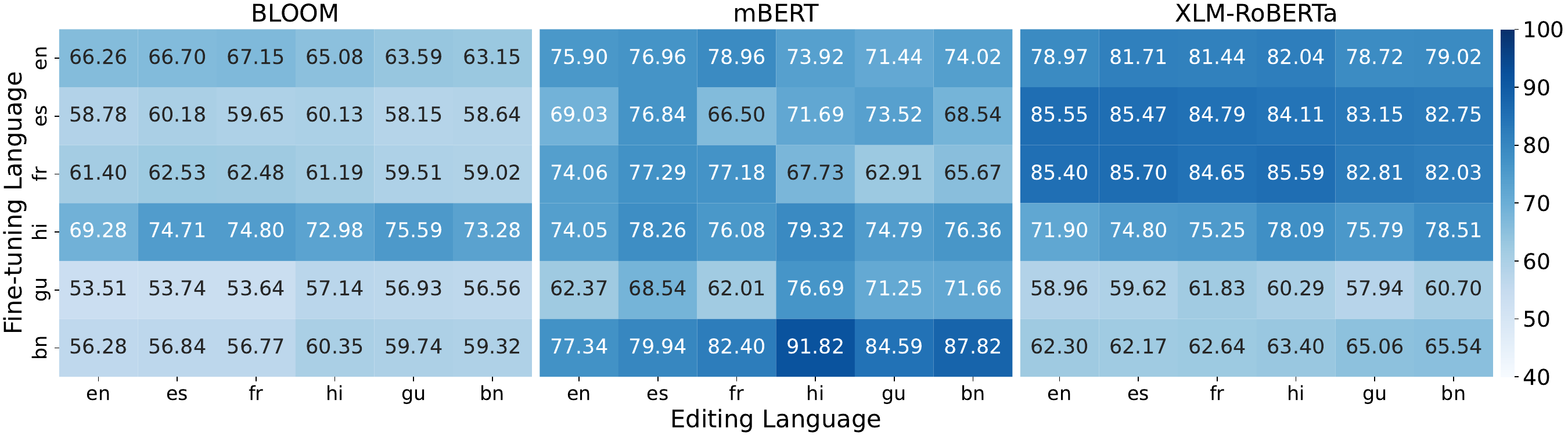}
\end{center}
\caption{The figure illustrates $G_S$ given the editing language (x-axis) and fine-tuning languages (y-axis) for all the three models \texttt{BLOOM} (left), \texttt{mBERT} (middle) and \texttt{XLM-RoBERTa} (right) when edited using \textbf{KE}.}
\label{fig:ke_wl}
\end{figure*}

\begin{figure}
\begin{center}
\includegraphics[width=0.8\linewidth]{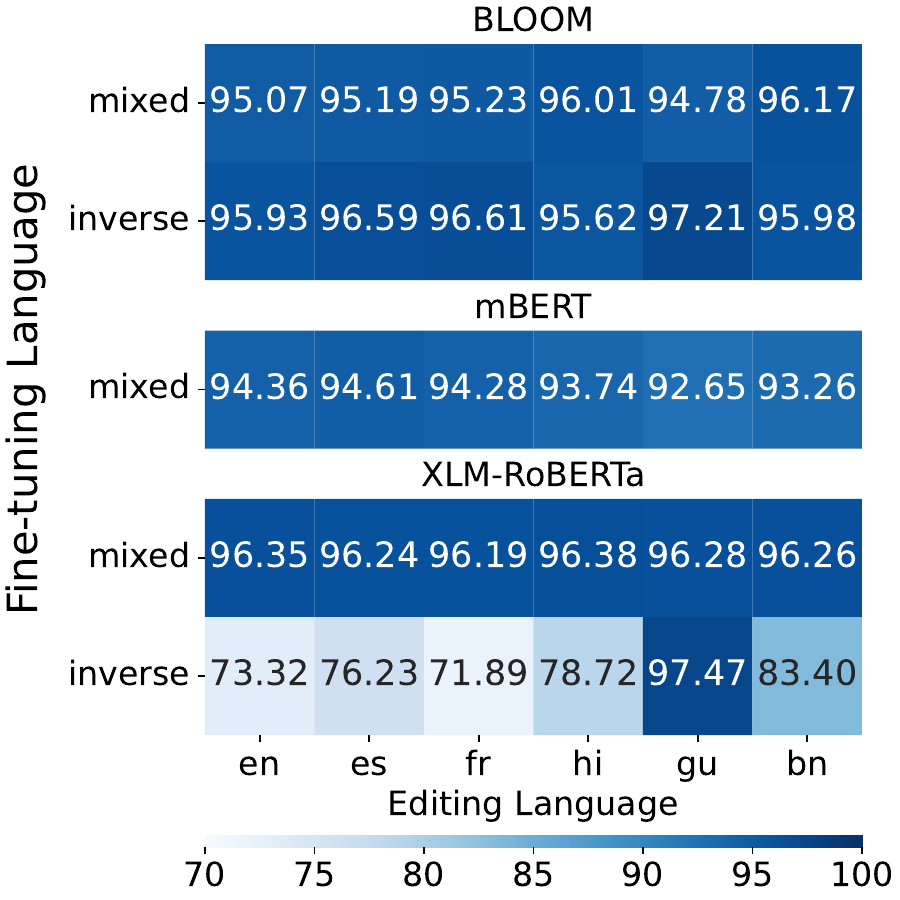}
\end{center}
\caption{The figure illustrates $S_S$ given the editing language (x-axis) and fine-tuning datasets (y-axis) for all the three models \texttt{BLOOM} (top), \texttt{mBERT} (middle) and \texttt{XLM-RoBERTa} (right) when edited using \textbf{KE}.}
\label{fig:ke_wl_mi_loc}
\end{figure}

\begin{figure}
\begin{center}
\includegraphics[width=0.8\linewidth]{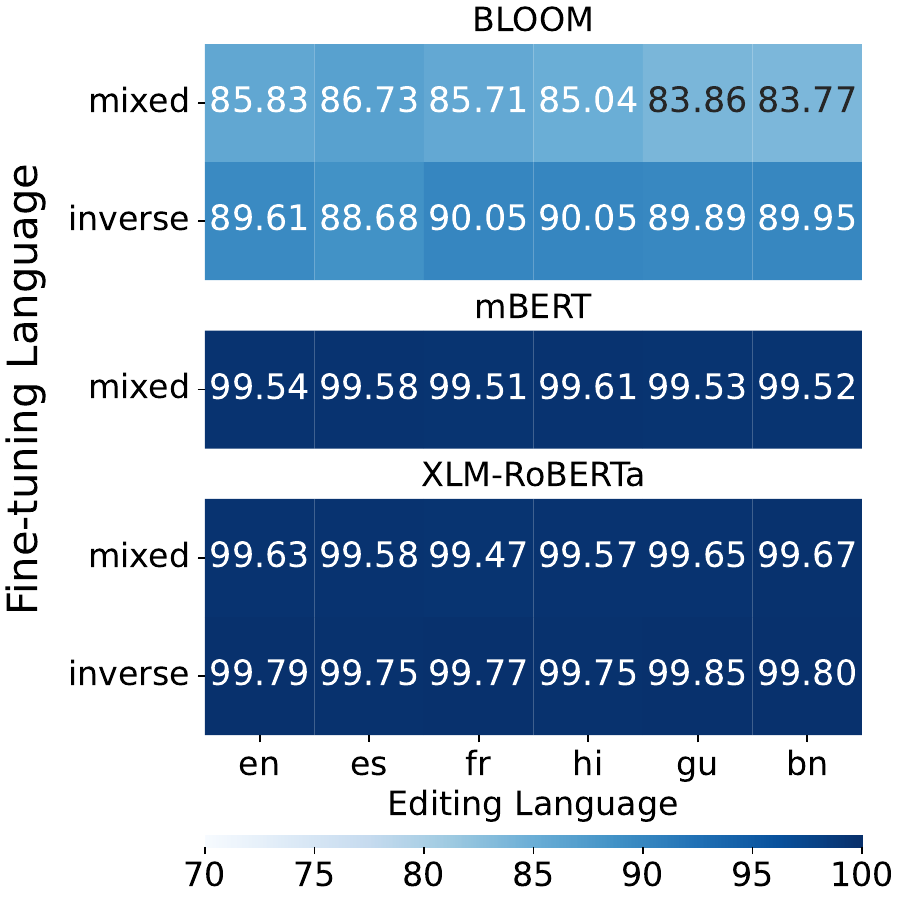}
\end{center}
\caption{The figure illustrates $S_S$ given the editing language (x-axis) and fine-tuning datasets (y-axis) for all the three models \texttt{BLOOM} (top), \texttt{mBERT} (middle) and \texttt{XLM-RoBERTa} (right) when edited using \textbf{FT}.}
\label{fig:ft_wl_mi_loc}
\end{figure}

\begin{figure*}
\begin{center}
\includegraphics[width=\linewidth]{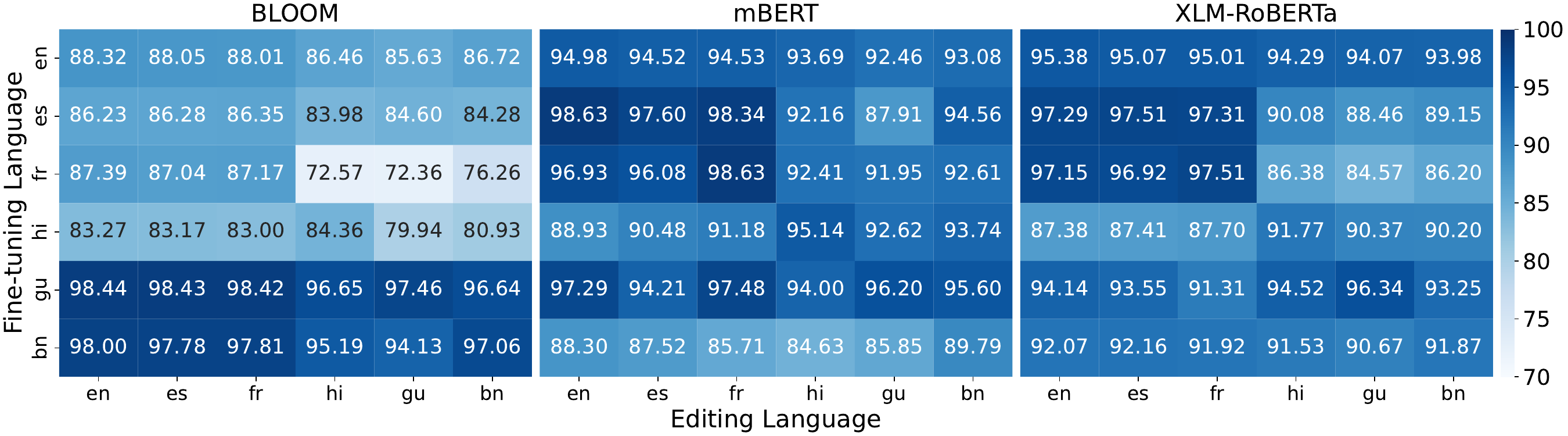}
\end{center}
\caption{The figure illustrates $S_S$ given the editing language (x-axis) and fine-tuning languages (y-axis) for all the three models \texttt{BLOOM} (left), \texttt{mBERT} (middle) and \texttt{XLM-RoBERTa} (right) when edited using \textbf{KE}.}
\label{fig:ke_wl_loc}
\end{figure*}

\subsection{Dataset}
\label{Appendix:dataset}
The complete dataset statistic regarding the cross-lingual dataset and Average Lengths (AL) for encoder-only and decoder-only models are shown in Table \ref{stasdataset}. We considered the samples overlapping in all six languages (not including mixed and inverse) from the train, validation, and test splits. Table \ref{dataset_inv_bloom} and \ref{dataset_inv_xlm} report the inverse proportion of languages for BLOOM and XLM-RoBERTa.
\subsection{Quality Assessment of Translations}
\label{sec:annotation}
We randomly selected 150 instances from the English-FEVER dataset \citep{fever} and the corresponding translations and then assigned them to the human annotators. There were two annotators per language; each was a native speaker of the language assigned to them and proficient in English. We recruited language experts who voluntarily helped in the annotation process without pay. 

Table~\ref{tab:annotation} shows the individual annotation accuracy and inter-annotation agreement (IAA). In the table, the IAA column represents scores computed from Cohen's Kappa coefficient, computed between two annotators for the respective language. While computing the IAA, annotators verified that the translated sentences were syntactically and semantically correct (No code-switching or code-mixing was allowed). Considering the Relaxed-IAA (R-IAA), code-mixed and code-switched transitions were assumed to be relaxed and surpassed (Correct semantics were verified). Further, $acc_{a1}$ and $acc_{a2}$ represent the accuracy\footnote{We have computed average accuracy as the ratio of correct translations annotated with the total number of instances.} of annotators one and two with strict instructions. Lastly, R-$acc_{a1}$ and R-$acc_{a2}$ represent the accuracy with the relaxed instructions from both annotators. Accuracy for individual annotators was over 80 percent in all the cases.

% IAA is computed in two configurations: \textbf{\textit{(1)}} Standard IAA, aka Cohen's Kappa Score, and \textbf{\textit{(2)}} IAA-Relaxed (Cohen's Kappa on relaxed annotations). The standard IAA is evaluated with the instructions that the translations should be syntactically and semantically correct, including translations for numeric digits. However, the IAA-Relaxed is evaluated with the instructions of correct semantics, and it is a relaxed version of the standard IAA. 
% 
% Overall, the \textit{Latin} script has shown better inter-annotator agreement and average annotator accuracy than the \textit{Indic} script languages. 

\begin{table*}
\centering
\begin{tabular}{ccccccccc} % l r c p{2cm}
\hline
\textbf{Language} & \textbf{IAA} & \textbf{R-IAA} & \textbf{$acc_{a1}$} & \textbf{$acc_{a2}$} & \textbf{Avg. Acc.} & \textbf{R-$acc_{a1}$} & \textbf{R-$acc_{a2}$} & \textbf{R-Avg. Acc.} \\ \hline
French   & 67.00 & 80.00 & 88.67 & 94    & 91.33 & 92.00    & 93.33 & 92.66 \\
Spanish  & 66.00 & 74.00 & 76.67 & 84.67 & 80.67 & 87.33 & 90.00    & 88.66 \\
Hindi    & 63.00 & 85.00 & 75.33 & 76.67 & 76.00    & 93.33 & 92.67 & 93.00     \\
Bengali  & 70.00 & 76.00 & 80.67 & 80.67 & 80.67 & 92.67 & 92.00    & 92.335 \\
Gujarati & 56.00 & 74.00 & 66.67 & 59.33 & 63.00    & 74.00    & 73.33 & 73.66\\ \hline
\textbf{Average} & 64.4 & 77.8 & 77.60 & 79.07 & 78.33 & 87.87 & 88.27 & 88.07 \\  \hline
\end{tabular}
\caption{\label{tab:annotation}
Inter-Annotator Agreement (IAA), Relaxed-IAA, and average accuracy per language the annotators assign for both standard and relaxed configurations (Reported numbers are percentages over 150 instances). In our experiments, two annotators represented as $a1$ and $a2$ were asked to annotate the correct translations. Standard accuracy per language by annotator is represented with $acc_{a1}$ and $acc_{a2}$, whereas relaxed accuracy is denoted with R-$acc_{a1}$ and R-$acc_{a2}$ for annotators one and two.
}
\end{table*}
\subsection{Model Editing Techniques}
\label{relatedMEtechniques}
Table \ref{recent_workstable} reports the 24 editing techniques introduced in top venues over the recent years. The techniques are classified into different editing approaches. From the literature review, the editing techniques have gained popularity and trends to become a focused problem for the future never-aging LLMs. Figure \ref{fig:ke_wl} shows the average $G_S$ for all three models for \textbf{KE}. Furthermore, Figure \ref{fig:ke_wl_mi} shows the mixed and inverse proportion results for the \textbf{KE} and \textbf{FT}. Similarly, Figures \ref{fig:ke_wl_loc} and \ref{fig:ft_wl_loc} show the average $S_S$ for all three models for \textbf{KE} and \textbf{FT}. Furthermore, Figure \ref{fig:ke_wl_mi_loc} and \ref{fig:ft_wl_mi_loc} shows the mixed and inverse proportion results for the \textbf{KE} and \textbf{FT}.
\subsection{Implementation Details}
We utilized the \citet{mend}'s implementation of \textbf{MEND}, \textbf{KE}, and \textbf{FT}. We used the default hyperparameters to fine-tune the base model and the MLPs as specified in MEND's implementation. We edit one instance per batch. 
For all 69 configurations with Language Pairs x Models x METs, a total of 9,936 experiments were performed. From the tables indexed in \ref{index_table}, one experiment is computed as $G_S$ and $S_S$ for one configuration, say, in Table \ref{tab:mend_bloom-560M_en}, for IL, when $x$ is \textbf{en}, and $x'$ is en for both $G_S$ and $S_S$. Similarly, for one set of layers (36 values), there are a total of 4 sets and 69 configurations, which sums to 36 x 4 x 69 = 9,936 experiments. 
% \subsection{Experiments}
% \label{sec:exp_all}

\subsection{MEND}
\label{sec:mend_all}
\subsubsection{BLOOM}
\label{sec:mend_bloom}
Tables \ref{tab:mend_bloom-560M_en}, \ref{tab:mend_bloom-560M_fr}, \ref{tab:mend_bloom-560M_es}, \ref{tab:mend_bloom-560M_hi}, \ref{tab:mend_bloom-560M_gu}, \ref{tab:mend_bloom-560M_bn}, \ref{tab:mend_bloom-560M_mixed}, and \ref{tab:mend_bloom-560M_inverse} shows the experiments on BLOOM when fine-tuned on en, fr, es, hi, gu, bn, mixed, and inverse, respectively using \textbf{MEND}.

\subsubsection{mBERT}
\label{sec:mend_mBERT}
Tables \ref{tab:mend_mBERT_en}, \ref{tab:mend_mBERT_fr}, \ref{tab:mend_mBERT_es}, \ref{tab:mend_mBERT_hi}, \ref{tab:mend_mBERT_gu}, \ref{tab:mend_mBERT_bn}, and \ref{tab:mend_mBERT_mixed}, shows the experiments on mBERT when fine-tuned on en, fr, es, hi, gu, bn, and mixed, respectively using \textbf{MEND}.

\subsubsection{XLM-RoBERTa}
\label{sec:mend_xlm}
Tables \ref{tab:mend_XLM-RoBERTa_en}, \ref{tab:mend_XLM-RoBERTa_fr}, \ref{tab:mend_XLM-RoBERTa_es}, \ref{tab:mend_XLM-RoBERTa_hi}, \ref{tab:mend_XLM-RoBERTa_gu}, \ref{tab:mend_XLM-RoBERTa_bn}, \ref{tab:mend_XLM-RoBERTa_mixed}, and \ref{tab:mend_XLM-RoBERTa_inverse}, shows the experiments on XLM-RoBERTa when fine-tuned on en, fr, es, hi, gu, bn, and mixed, respectively using \textbf{MEND}.

\subsection{KE}
\label{sec:ke_all}

\subsubsection{BLOOM}
\label{sec:ke_bloom}
Tables \ref{tab:ke_bloom-560M_en}, \ref{tab:ke_bloom-560M_fr}, \ref{tab:ke_bloom-560M_es}, \ref{tab:ke_bloom-560M_hi}, \ref{tab:ke_bloom-560M_gu}, \ref{tab:ke_bloom-560M_bn}, \ref{tab:ke_bloom-560M_mixed}, and \ref{tab:ke_bloom-560M_inverse} shows the experiments on BLOOM when fine-tuned on en, fr, es, hi, gu, bn, mixed, and inverse, respectively using \textbf{KE}.

\subsubsection{mBERT}
\label{sec:ke_mBERT}
Tables \ref{tab:ke_mBERT_en}, \ref{tab:ke_mBERT_fr}, \ref{tab:ke_mBERT_es}, \ref{tab:ke_mBERT_hi}, \ref{tab:ke_mBERT_gu}, \ref{tab:ke_mBERT_bn}, and \ref{tab:ke_mBERT_mixed}, shows the experiments on mBERT when fine-tuned on en, fr, es, hi, gu, bn, and mixed, respectively using \textbf{ke}.

\subsubsection{XLM-RoBERTa}
\label{sec:ke_xlm}
Tables \ref{tab:ke_XLM-RoBERTa_en}, \ref{tab:ke_XLM-RoBERTa_fr}, \ref{tab:ke_XLM-RoBERTa_es}, \ref{tab:ke_XLM-RoBERTa_hi}, \ref{tab:ke_XLM-RoBERTa_gu}, \ref{tab:ke_XLM-RoBERTa_bn}, \ref{tab:ke_XLM-RoBERTa_mixed}, and \ref{tab:ke_XLM-RoBERTa_inverse}, shows the experiments on XLM-RoBERTa when fine-tuned on en, fr, es, hi, gu, bn, and mixed, respectively using \textbf{MEND}.

\subsection{FT}
\label{sec:ft_all}
\subsubsection{BLOOM}
\label{sec:ft_bloom}
Tables \ref{tab:ft_bloom-560M_en}, \ref{tab:ft_bloom-560M_fr}, \ref{tab:ft_bloom-560M_es}, \ref{tab:ft_bloom-560M_hi}, \ref{tab:ft_bloom-560M_gu}, \ref{tab:ft_bloom-560M_bn}, \ref{tab:ft_bloom-560M_mixed}, and \ref{tab:ft_bloom-560M_inverse} shows the experiments on BLOOM when fine-tuned on en, fr, es, hi, gu, bn, mixed, and inverse, respectively using \textbf{FT}.

\subsubsection{mBERT}
\label{sec:ft_mBERT}
Tables \ref{tab:ke_mBERT_en}, \ref{tab:ke_mBERT_fr}, \ref{tab:ke_mBERT_es}, \ref{tab:ke_mBERT_hi}, \ref{tab:ke_mBERT_gu}, \ref{tab:ke_mBERT_bn}, and \ref{tab:ke_mBERT_mixed}, shows the experiments on mBERT when fine-tuned on en, fr, es, hi, gu, bn, and mixed, respectively using \textbf{FT}.

\subsubsection{XLM-RoBERTa}
\label{sec:ft_xlm}
Tables \ref{tab:ft_XLM-RoBERTa_en}, \ref{tab:ft_XLM-RoBERTa_fr}, \ref{tab:ft_XLM-RoBERTa_es}, \ref{tab:ft_XLM-RoBERTa_hi}, \ref{tab:ft_XLM-RoBERTa_gu}, \ref{tab:ft_XLM-RoBERTa_bn}, \ref{tab:ft_XLM-RoBERTa_mixed}, and \ref{tab:ft_XLM-RoBERTa_inverse}, shows the experiments on XLM-RoBERTa when fine-tuned on en, fr, es, hi, gu, bn, and mixed, respectively using \textbf{FT}.

All 69 configurations for ME techniques, models, and languages are indexed to Table \ref{index_table}. The normalized $G_S$ for \textbf{KE} and \textbf{FT} are shown in Figure \ref{fig:ke_wl} and \ref{fig:ft_wl}, respectively. Furthermore, Figures \ref{fig:ke_wl_mi} and \ref{fig:ft_wl_mi} show the normalized $G_S$ for \textbf{KE} and \textbf{FT} for mixed and inverse configurations, respectively using \textbf{MEND}.

 \begin{table*}[htpb]
\centering
\begin{tabular}{p{5cm}lllll}
\hline
\textbf{Technique} & \textbf{Venue} & \textbf{On Arxiv}  & \textbf{Cit.} & \textbf{Technique} & \textbf{Code}\\
\hline
ENNs \citep{enn}         & ICLR 20'           & Apr 01, 2020      & 76                 & LE  & \href{https://github.com/xtinkt/editable}{Y}      \\
KE \citep{knowledgeediting}   & EMNLP 21'          & Apr 16, 2021      & 75                 & HN   &  \href{https://github.com/nicola-decao/KnowledgeEditor}{Y}     \\
KN  \citep{knowledgeneurons} & ACL Proceeding 22’ & Apr 18, 2021      & 75                 & HN   &  \href{https://github.com/EleutherAI/knowledge-neurons}{Y}     \\
CuQA  \citep{CuQA}                         & ACL Proceeding 22’ & Apr 22, 2021      & 1                  &  LE &    \href{https://aclanthology.org/2022.findings-acl.37/}{Y}      \\
MEND  \citep{mend}                         & ICLR 22’           & Oct 21, 2021      & 73                 &  HN  &    \href{https://github.com/eric-mitchell/mend}{-}     \\
SLAG \citep{slag}                          & Arxiv              & Nov 26, 2021      & 21                 &  LE  &    \href{https://github.com/peterbhase/SLAG-Belief-Updating}{Y}     \\
Editing-classifier \citep{edit_class}      & NeurIPS 21’        & Dec 01, 2021      & 30                 &    LE    &  \href{https://github.com/MadryLab/EditingClassifiers}{Y}   \\
\hline
\end{tabular}
\caption{\label{recent_workstable}
Recent works in METs. Note: Citations were last reported on April 11, 2023. Here, PP stands for `Parameter Preserving,' HN for `HyperNetworks,' and LE for `Localized Edits.'
}
\end{table*}

 \begin{table*}[htpb]
\centering
\begin{tabular}{ccccccc} % l r c p{2cm}
\hline
\textbf{Lang} &
  \textbf{AL$_\alpha$} &
  \textbf{AL$_\beta$} &
  \textbf{AL$_\gamma$} &
  \textbf{Train} &
  \textbf{TFR} &
  \textbf{VFR} \\ \hline
en            & 11.25 & 10.67 & 11.87 & 104966 & 10.9998 & 10.5003 \\
hi            & 14.4  & 18.04 & 15.69 & 103191 & 10.691  & 10.2668 \\
es            & 12.25 & 12.53 & 14.07 & 104965 & 10.8479 & 10.3747 \\
fr            & 10.5  & 10.6  & 12.79 & 104966 & 10.8479 & 10.3529 \\
bn            & 13.58 & 20.72 & 17.61 & 104966 & 10.8479 & 10.3747 \\
gu            & 15.93 & 23.86 & 18.07 & 104966 & 10.8479 & 10.3747 \\
mix           & 11.25 & 10.67 & 11.25 & 102922 & 10.8633 & 10.4186 \\
$Inv_{bloom}$ & 11.25 & -     & -     & 104504 & 10.8437 & 10.3747 \\
$Inv_{xlm}$   & -     & -     & 11.95 & 104966 & 10.8483 & 10.3747 \\ \hline
\end{tabular}
\caption{\label{stasdataset}
Dataset statistics in different languages. Note TFR and VFR are the average length of training-filtered and validation filtered rephrases, respectively. $Inv_{bloom}$ and $Inv_{xlm}$ are the inverse proportion of \texttt{BLOOM} and \texttt{XLM-RoBERTa}. We do not include the inverse proportion of \texttt{mBERT}. Lastly, in all the languages, the size of validation and test remains 10444 and 1193, respectively. 
}
\end{table*}

 \begin{table*}[htpb]
\centering
\begin{tabular}{llllll} \hline
\textbf{Language} & \textbf{Size} (x$10^{10}$)         & \textbf{Proportion (In \%)}           & \textbf{Inv. Proportion}    & \textbf{Train}  & \textbf{Test}  \\ \hline
en  & 48.50 & 53.13 & 1.88     & 230    & 23    \\
fr   & 20.82 & 22.82 & 4.38     & 536    & 53    \\
es  & 17.50 & 19.18 & 5.21     & 638    & 63    \\
hi    & 2.46  & 2.7 & 37.07    & 4534   & 451   \\
gu & 1.86  &2.04 & 49.05    & 6000   & 597   \\
bn  & 0.12   & 0.13    & 760.61   & 93029  & 9256  \\ \hline
Total    & 91.27 & 1    & 858.21   & 104967 & 10443 \\ \hline
\end{tabular}
\caption{\label{dataset_inv_bloom}
Statistics of inverse proportion dataset for \texttt{BLOOM}. The dataset is prepared while considering the portion of languages at the pre-training stage. Proportion is shown in percentage over the six languages. 
}
\end{table*}
 \begin{table*}[htpb]
\centering
\begin{tabular}{llllll} \hline
\textbf{Language} & \textbf{Size} (x$10^{3}$)        & \textbf{Proportion (In \%)}           & \textbf{Inv. Proportion}    & \textbf{Train}  & \textbf{Test}  \\ \hline
en    & 55.6 & 72.09 & 1.39 & 191    & 19    \\
fr    & 9.78  & 12.68   & 7.89 & 1089   & 108   \\
es    & 9.37  & 12.15    & 8.23 & 1136   & 113   \\
hi    & 1.71  & 2.22   & 44.98 & 6209   & 618   \\
gu    & 0.53   & 0.68 & 146.93 & 20282  & 2018  \\ 
bn    & 0.14   & 0.18  & 551.01 & 76059  & 7568  \\ \hline
Total & 77.14 & 1              & 760.44 & 104966 & 10444 \\ \hline
\end{tabular}
\caption{\label{dataset_inv_xlm}
Statistics of inverse proportion dataset for \texttt{xlm-roberta}. The dataset is prepared while considering the portion of languages at the pre-training stage. Proportion is shown in percentage over the six languages. 
}
\end{table*}

%%%%%%%%%%%%%%%%%%%%%%%% TABLE START MEND-BLOOM %%%%%%%%%%%%%%%%%%%%%%%%%%%% 
%%%%%%%%%%%%%%%%%%%%%%%% MEND-BLOOM-en %%%%%%%%%%%%%%%%%%%%%%%%%%%% 
\begin{table*}[htpb]
\centering

\resizebox{\linewidth}{!}{%
\begin{tabular}{>{\hspace{0pt}}m{0.020\linewidth}>{\hspace{0pt}}m{0.045\linewidth}|>{\centering\hspace{0pt}}m{0.069\linewidth}>{\centering\hspace{0pt}}m{0.069\linewidth}>{\centering\hspace{0pt}}m{0.069\linewidth}>{\centering\hspace{0pt}}m{0.069\linewidth}>{\centering\hspace{0pt}}m{0.069\linewidth}>{\centering\hspace{0pt}}m{0.069\linewidth}|>{\centering\hspace{0pt}}m{0.058\linewidth}>{\centering\hspace{0pt}}m{0.058\linewidth}>{\centering\hspace{0pt}}m{0.058\linewidth}>{\centering\hspace{0pt}}m{0.069\linewidth}>{\centering\hspace{0pt}}m{0.069\linewidth}>{\centering\arraybackslash\hspace{0pt}}m{0.069\linewidth}} 
\hline
 &  & \multicolumn{6}{>{\centering\hspace{0pt}}m{0.39\linewidth}|}{$G_S$  ($x'$) $\rightarrow$} & \multicolumn{6}{>{\centering\arraybackslash\hspace{0pt}}m{0.39\linewidth}}{$S_S$ ($\hat{x}$) $\rightarrow$} \\ 
\hline\multicolumn{1}{>{\centering\hspace{0pt}}m{0.05\linewidth}}{\textbf{Set}} & \textbf{$x$} $\downarrow$ & \textbf{en} & \textbf{fr} & \textbf{es} & \textbf{hi} & \textbf{gu} & \textbf{bn} & \textbf{en} & \textbf{fr} & \textbf{es} & \textbf{hi} & \textbf{gu} & \textbf{bn} \\ \hline\multicolumn{1}{>{\centering\hspace{0pt}}m{0.05\linewidth}}{\textbf{IL}}& en & \cellcolor[HTML]{4C99DB}\textbf{87.93}& \cellcolor[HTML]{6BAAE1}79.8& \cellcolor[HTML]{67A8E1}80.72& \cellcolor[HTML]{B4D4F0}59.93& \cellcolor[HTML]{DFEDF8}48.37& \cellcolor[HTML]{BBD8F1}58.26& \cellcolor[HTML]{3A8ED8}96.5& \cellcolor[HTML]{368CD7}\textbf{96.96}& \cellcolor[HTML]{388DD7}96.75& \cellcolor[HTML]{83B8E6}86.61& \cellcolor[HTML]{549DDD}92.98& \cellcolor[HTML]{7BB4E5}87.66 \\ 
& fr & \cellcolor[HTML]{77B2E4}76.36& \cellcolor[HTML]{4E9ADC}\textbf{87.43}& \cellcolor[HTML]{63A6E0}81.81& \cellcolor[HTML]{BBD8F1}58.26& \cellcolor[HTML]{DCEBF8}49.29& \cellcolor[HTML]{C0DBF2}56.92& \cellcolor[HTML]{3189D6}\textbf{97.65}& \cellcolor[HTML]{368CD7}96.94& \cellcolor[HTML]{358BD7}97.17& \cellcolor[HTML]{78B2E4}88.14& \cellcolor[HTML]{4796DA}94.74& \cellcolor[HTML]{72AEE3}88.92 \\
\hline 

\end{tabular}
}
\caption{\label{tab:mend_bloom-560M_en}
The table represents the $G_S$ and $S_S$ using \textbf{MEND} over fine-tuned \texttt{BLOOM} on the fever `\textbf{en}' dataset. 
}
\end{table*} 
%%%%%%%%%%%%%%%%%%%%%%%% MEND-BLOOM-fr %%%%%%%%%%%%%%%%%%%%%%%%%%%% 
\begin{table*}[htpb]
\centering

\resizebox{\linewidth}{!}{%
\begin{tabular}{>{\hspace{0pt}}m{0.020\linewidth}>{\hspace{0pt}}m{0.045\linewidth}|>{\centering\hspace{0pt}}m{0.069\linewidth}>{\centering\hspace{0pt}}m{0.069\linewidth}>{\centering\hspace{0pt}}m{0.069\linewidth}>{\centering\hspace{0pt}}m{0.069\linewidth}>{\centering\hspace{0pt}}m{0.069\linewidth}>{\centering\hspace{0pt}}m{0.069\linewidth}|>{\centering\hspace{0pt}}m{0.058\linewidth}>{\centering\hspace{0pt}}m{0.058\linewidth}>{\centering\hspace{0pt}}m{0.058\linewidth}>{\centering\hspace{0pt}}m{0.069\linewidth}>{\centering\hspace{0pt}}m{0.069\linewidth}>{\centering\arraybackslash\hspace{0pt}}m{0.069\linewidth}} 
\hline
 &  & \multicolumn{6}{>{\centering\hspace{0pt}}m{0.39\linewidth}|}{$G_S$  ($x'$) $\rightarrow$} & \multicolumn{6}{>{\centering\arraybackslash\hspace{0pt}}m{0.39\linewidth}}{$S_S$ ($\hat{x}$) $\rightarrow$} \\ 
\hline\multicolumn{1}{>{\centering\hspace{0pt}}m{0.05\linewidth}}{\textbf{Set}} & \textbf{$x$} $\downarrow$ & \textbf{en} & \textbf{fr} & \textbf{es} & \textbf{hi} & \textbf{gu} & \textbf{bn} & \textbf{en} & \textbf{fr} & \textbf{es} & \textbf{hi} & \textbf{gu} & \textbf{bn} \\ \hline\multicolumn{1}{>{\centering\hspace{0pt}}m{0.05\linewidth}}{\textbf{IL}}& en & \cellcolor[HTML]{69A9E1}\textbf{80.22}& \cellcolor[HTML]{8ABCE7}71.42& \cellcolor[HTML]{7CB4E5}75.11& \cellcolor[HTML]{D2E5F6}51.89& \cellcolor[HTML]{C4DDF3}55.66& \cellcolor[HTML]{C9E0F4}54.4& \cellcolor[HTML]{569EDD}92.69& \cellcolor[HTML]{529CDD}\textbf{93.17}& \cellcolor[HTML]{5EA3DF}91.53& \cellcolor[HTML]{FFFFFF}69.09& \cellcolor[HTML]{FFFFFF}62.01& \cellcolor[HTML]{FFFFFF}63.77 \\ 
& fr & \cellcolor[HTML]{88BBE7}71.75& \cellcolor[HTML]{6CABE2}\textbf{79.46}& \cellcolor[HTML]{7DB5E5}74.94& \cellcolor[HTML]{D0E4F5}52.56& \cellcolor[HTML]{C3DCF3}56.08& \cellcolor[HTML]{C8E0F4}54.57& \cellcolor[HTML]{569EDD}92.69& \cellcolor[HTML]{509BDC}\textbf{93.46}& \cellcolor[HTML]{5EA3DF}91.53& \cellcolor[HTML]{FFFFFF}64.48& \cellcolor[HTML]{FFFFFF}58.53& \cellcolor[HTML]{FFFFFF}59.03 \\
\hline 

\end{tabular}
}
\caption{\label{tab:mend_bloom-560M_fr}
The table represents the $G_S$ and $S_S$ using \textbf{MEND} over fine-tuned \texttt{BLOOM} on the fever `\textbf{fr}' dataset. 
}
\end{table*} 
%%%%%%%%%%%%%%%%%%%%%%%% MEND-BLOOM-es %%%%%%%%%%%%%%%%%%%%%%%%%%%% 
\begin{table*}[htpb]
\centering

\resizebox{\linewidth}{!}{%
\begin{tabular}{>{\hspace{0pt}}m{0.020\linewidth}>{\hspace{0pt}}m{0.045\linewidth}|>{\centering\hspace{0pt}}m{0.069\linewidth}>{\centering\hspace{0pt}}m{0.069\linewidth}>{\centering\hspace{0pt}}m{0.069\linewidth}>{\centering\hspace{0pt}}m{0.069\linewidth}>{\centering\hspace{0pt}}m{0.069\linewidth}>{\centering\hspace{0pt}}m{0.069\linewidth}|>{\centering\hspace{0pt}}m{0.058\linewidth}>{\centering\hspace{0pt}}m{0.058\linewidth}>{\centering\hspace{0pt}}m{0.058\linewidth}>{\centering\hspace{0pt}}m{0.069\linewidth}>{\centering\hspace{0pt}}m{0.069\linewidth}>{\centering\arraybackslash\hspace{0pt}}m{0.069\linewidth}} 
\hline
 &  & \multicolumn{6}{>{\centering\hspace{0pt}}m{0.39\linewidth}|}{$G_S$  ($x'$) $\rightarrow$} & \multicolumn{6}{>{\centering\arraybackslash\hspace{0pt}}m{0.39\linewidth}}{$S_S$ ($\hat{x}$) $\rightarrow$} \\ 
\hline\multicolumn{1}{>{\centering\hspace{0pt}}m{0.05\linewidth}}{\textbf{Set}} & \textbf{$x$} $\downarrow$ & \textbf{en} & \textbf{fr} & \textbf{es} & \textbf{hi} & \textbf{gu} & \textbf{bn} & \textbf{en} & \textbf{fr} & \textbf{es} & \textbf{hi} & \textbf{gu} & \textbf{bn} \\ \hline\multicolumn{1}{>{\centering\hspace{0pt}}m{0.05\linewidth}}{\textbf{IL}}& en & \cellcolor[HTML]{5DA3DF}\textbf{83.4}& \cellcolor[HTML]{84B9E6}72.84& \cellcolor[HTML]{7EB5E5}74.52& \cellcolor[HTML]{FAFCFE}41.24& \cellcolor[HTML]{E9F2FA}45.85& \cellcolor[HTML]{D2E5F6}52.05& \cellcolor[HTML]{378DD7}\textbf{96.84}& \cellcolor[HTML]{4193D9}95.45& \cellcolor[HTML]{4A98DB}94.26& \cellcolor[HTML]{F0F6FC}71.96& \cellcolor[HTML]{A5CCED}81.98& \cellcolor[HTML]{FFFFFF}64.38 \\ 
& fr & \cellcolor[HTML]{89BCE7}71.58& \cellcolor[HTML]{73AFE3}\textbf{77.62}& \cellcolor[HTML]{7DB5E5}74.94& \cellcolor[HTML]{FBFCFE}40.99& \cellcolor[HTML]{E9F2FA}45.68& \cellcolor[HTML]{D1E5F6}52.22& \cellcolor[HTML]{348BD7}\textbf{97.21}& \cellcolor[HTML]{4394DA}95.22& \cellcolor[HTML]{4D99DB}93.88& \cellcolor[HTML]{F4F8FC}71.46& \cellcolor[HTML]{A6CCED}81.96& \cellcolor[HTML]{FFFFFF}64.65 \\
\hline 

\end{tabular}
}
\caption{\label{tab:mend_bloom-560M_es}
The table represents the $G_S$ and $S_S$ using \textbf{MEND} over fine-tuned \texttt{BLOOM} on the fever `\textbf{es}' dataset. 
}
\end{table*} 
%%%%%%%%%%%%%%%%%%%%%%%% MEND-BLOOM-hi %%%%%%%%%%%%%%%%%%%%%%%%%%%% 
\begin{table*}[htpb]
\centering

\resizebox{\linewidth}{!}{%
\begin{tabular}{>{\hspace{0pt}}m{0.020\linewidth}>{\hspace{0pt}}m{0.045\linewidth}|>{\centering\hspace{0pt}}m{0.069\linewidth}>{\centering\hspace{0pt}}m{0.069\linewidth}>{\centering\hspace{0pt}}m{0.069\linewidth}>{\centering\hspace{0pt}}m{0.069\linewidth}>{\centering\hspace{0pt}}m{0.069\linewidth}>{\centering\hspace{0pt}}m{0.069\linewidth}|>{\centering\hspace{0pt}}m{0.058\linewidth}>{\centering\hspace{0pt}}m{0.058\linewidth}>{\centering\hspace{0pt}}m{0.058\linewidth}>{\centering\hspace{0pt}}m{0.069\linewidth}>{\centering\hspace{0pt}}m{0.069\linewidth}>{\centering\arraybackslash\hspace{0pt}}m{0.069\linewidth}} 
\hline
 &  & \multicolumn{6}{>{\centering\hspace{0pt}}m{0.39\linewidth}|}{$G_S$  ($x'$) $\rightarrow$} & \multicolumn{6}{>{\centering\arraybackslash\hspace{0pt}}m{0.39\linewidth}}{$S_S$ ($\hat{x}$) $\rightarrow$} \\ 
\hline\multicolumn{1}{>{\centering\hspace{0pt}}m{0.05\linewidth}}{\textbf{Set}} & \textbf{$x$} $\downarrow$ & \textbf{en} & \textbf{fr} & \textbf{es} & \textbf{hi} & \textbf{gu} & \textbf{bn} & \textbf{en} & \textbf{fr} & \textbf{es} & \textbf{hi} & \textbf{gu} & \textbf{bn} \\ \hline\multicolumn{1}{>{\centering\hspace{0pt}}m{0.05\linewidth}}{\textbf{IL}}& en & \cellcolor[HTML]{61A5DF}\textbf{82.31}& \cellcolor[HTML]{6EACE2}78.79& \cellcolor[HTML]{70ADE2}78.37& \cellcolor[HTML]{B5D5F0}59.77& \cellcolor[HTML]{CEE3F5}53.14& \cellcolor[HTML]{BBD8F1}58.26& \cellcolor[HTML]{FFFFFF}64.48& \cellcolor[HTML]{FFFFFF}68.42& \cellcolor[HTML]{FFFFFF}67.48& \cellcolor[HTML]{94C2E9}\textbf{84.37}& \cellcolor[HTML]{E0EDF8}74.1& \cellcolor[HTML]{E0EDF8}74.12 \\ 
& fr & \cellcolor[HTML]{5DA2DF}83.49& \cellcolor[HTML]{4D99DB}\textbf{87.85}& \cellcolor[HTML]{529CDC}86.5& \cellcolor[HTML]{ABCFEE}62.45& \cellcolor[HTML]{C1DBF2}56.66& \cellcolor[HTML]{B4D4F0}60.02& \cellcolor[HTML]{E3EFF9}73.68& \cellcolor[HTML]{FFFFFF}68.13& \cellcolor[HTML]{F2F7FC}71.71& \cellcolor[HTML]{79B2E4}\textbf{88.01}& \cellcolor[HTML]{BAD8F1}79.19& \cellcolor[HTML]{C2DCF3}78.14 \\
\hline 

\end{tabular}
}
\caption{\label{tab:mend_bloom-560M_hi}
The table represents the $G_S$ and $S_S$ using \textbf{MEND} over fine-tuned \texttt{BLOOM} on the fever `\textbf{hi}' dataset. 
}
\end{table*} 
%%%%%%%%%%%%%%%%%%%%%%%% MEND-BLOOM-gu %%%%%%%%%%%%%%%%%%%%%%%%%%%% 
\begin{table*}[htpb]
\centering

\resizebox{\linewidth}{!}{%
\begin{tabular}{>{\hspace{0pt}}m{0.020\linewidth}>{\hspace{0pt}}m{0.045\linewidth}|>{\centering\hspace{0pt}}m{0.069\linewidth}>{\centering\hspace{0pt}}m{0.069\linewidth}>{\centering\hspace{0pt}}m{0.069\linewidth}>{\centering\hspace{0pt}}m{0.069\linewidth}>{\centering\hspace{0pt}}m{0.069\linewidth}>{\centering\hspace{0pt}}m{0.069\linewidth}|>{\centering\hspace{0pt}}m{0.058\linewidth}>{\centering\hspace{0pt}}m{0.058\linewidth}>{\centering\hspace{0pt}}m{0.058\linewidth}>{\centering\hspace{0pt}}m{0.069\linewidth}>{\centering\hspace{0pt}}m{0.069\linewidth}>{\centering\arraybackslash\hspace{0pt}}m{0.069\linewidth}} 
\hline
 &  & \multicolumn{6}{>{\centering\hspace{0pt}}m{0.39\linewidth}|}{$G_S$  ($x'$) $\rightarrow$} & \multicolumn{6}{>{\centering\arraybackslash\hspace{0pt}}m{0.39\linewidth}}{$S_S$ ($\hat{x}$) $\rightarrow$} \\ 
\hline\multicolumn{1}{>{\centering\hspace{0pt}}m{0.05\linewidth}}{\textbf{Set}} & \textbf{$x$} $\downarrow$ & \textbf{en} & \textbf{fr} & \textbf{es} & \textbf{hi} & \textbf{gu} & \textbf{bn} & \textbf{en} & \textbf{fr} & \textbf{es} & \textbf{hi} & \textbf{gu} & \textbf{bn} \\ \hline\multicolumn{1}{>{\centering\hspace{0pt}}m{0.05\linewidth}}{\textbf{IL}}& en & \cellcolor[HTML]{C7DFF4}\textbf{54.99}& \cellcolor[HTML]{C7DFF4}54.82& \cellcolor[HTML]{C7DFF4}54.9& \cellcolor[HTML]{DCEBF8}49.29& \cellcolor[HTML]{DFECF8}48.53& \cellcolor[HTML]{CAE1F4}54.15& \cellcolor[HTML]{2080D3}\textbf{99.92}& \cellcolor[HTML]{2080D3}\textbf{99.92}& \cellcolor[HTML]{2080D3}99.81& \cellcolor[HTML]{4092D9}95.6& \cellcolor[HTML]{378DD7}96.81& \cellcolor[HTML]{4193D9}95.49 \\ 
& fr & \cellcolor[HTML]{C5DEF3}\textbf{55.57}& \cellcolor[HTML]{C5DEF3}55.41& \cellcolor[HTML]{C5DEF3}55.41& \cellcolor[HTML]{DAEAF7}49.79& \cellcolor[HTML]{E0EDF8}48.28& \cellcolor[HTML]{C9E0F4}54.4& \cellcolor[HTML]{2080D3}\textbf{99.92}& \cellcolor[HTML]{2080D3}99.89& \cellcolor[HTML]{2080D3}99.83& \cellcolor[HTML]{4092D9}95.66& \cellcolor[HTML]{368DD7}96.92& \cellcolor[HTML]{4092D9}95.68 \\
\hline 

\end{tabular}
}
\caption{\label{tab:mend_bloom-560M_gu}
The table represents the $G_S$ and $S_S$ using \textbf{MEND} over fine-tuned \texttt{BLOOM} on the fever `\textbf{gu}' dataset. 
}
\end{table*} 
%%%%%%%%%%%%%%%%%%%%%%%% MEND-BLOOM-bn %%%%%%%%%%%%%%%%%%%%%%%%%%%% 
\begin{table*}[htpb]
\centering

\resizebox{\linewidth}{!}{%
\begin{tabular}{>{\hspace{0pt}}m{0.020\linewidth}>{\hspace{0pt}}m{0.045\linewidth}|>{\centering\hspace{0pt}}m{0.069\linewidth}>{\centering\hspace{0pt}}m{0.069\linewidth}>{\centering\hspace{0pt}}m{0.069\linewidth}>{\centering\hspace{0pt}}m{0.069\linewidth}>{\centering\hspace{0pt}}m{0.069\linewidth}>{\centering\hspace{0pt}}m{0.069\linewidth}|>{\centering\hspace{0pt}}m{0.058\linewidth}>{\centering\hspace{0pt}}m{0.058\linewidth}>{\centering\hspace{0pt}}m{0.058\linewidth}>{\centering\hspace{0pt}}m{0.069\linewidth}>{\centering\hspace{0pt}}m{0.069\linewidth}>{\centering\arraybackslash\hspace{0pt}}m{0.069\linewidth}} 
\hline
 &  & \multicolumn{6}{>{\centering\hspace{0pt}}m{0.39\linewidth}|}{$G_S$  ($x'$) $\rightarrow$} & \multicolumn{6}{>{\centering\arraybackslash\hspace{0pt}}m{0.39\linewidth}}{$S_S$ ($\hat{x}$) $\rightarrow$} \\ 
\hline\multicolumn{1}{>{\centering\hspace{0pt}}m{0.05\linewidth}}{\textbf{Set}} & \textbf{$x$} $\downarrow$ & \textbf{en} & \textbf{fr} & \textbf{es} & \textbf{hi} & \textbf{gu} & \textbf{bn} & \textbf{en} & \textbf{fr} & \textbf{es} & \textbf{hi} & \textbf{gu} & \textbf{bn} \\ \hline\multicolumn{1}{>{\centering\hspace{0pt}}m{0.05\linewidth}}{\textbf{IL}}& en & \cellcolor[HTML]{C6DFF3}\textbf{55.07}& \cellcolor[HTML]{C7DFF4}54.9& \cellcolor[HTML]{CAE0F4}54.23& \cellcolor[HTML]{D6E8F7}50.8& \cellcolor[HTML]{E7F1FA}46.27& \cellcolor[HTML]{E3EFF9}47.44& \cellcolor[HTML]{2080D3}99.85& \cellcolor[HTML]{2080D3}99.87& \cellcolor[HTML]{2080D3}\textbf{99.89}& \cellcolor[HTML]{4997DB}94.36& \cellcolor[HTML]{2B86D5}98.49& \cellcolor[HTML]{358CD7}97.11 \\ 
& fr & \cellcolor[HTML]{C7DFF4}54.99& \cellcolor[HTML]{C6DEF3}\textbf{55.32}& \cellcolor[HTML]{C8DFF4}54.65& \cellcolor[HTML]{DEECF8}48.79& \cellcolor[HTML]{ECF4FB}45.01& \cellcolor[HTML]{E9F2FA}45.68& \cellcolor[HTML]{2080D3}99.96& \cellcolor[HTML]{2080D3}\textbf{99.98}& \cellcolor[HTML]{2080D3}\textbf{99.98}& \cellcolor[HTML]{4B98DB}94.13& \cellcolor[HTML]{2B86D5}98.45& \cellcolor[HTML]{348BD7}97.25 \\
\hline 

\end{tabular}
}
\caption{\label{tab:mend_bloom-560M_bn}
The table represents the $G_S$ and $S_S$ using \textbf{MEND} over fine-tuned \texttt{BLOOM} on the fever `\textbf{bn}' dataset. 
}
\end{table*} 
%%%%%%%%%%%%%%%%%%%%%%%% MEND-BLOOM-mixed %%%%%%%%%%%%%%%%%%%%%%%%%%%% 
\begin{table*}[htpb]
\centering

\resizebox{\linewidth}{!}{%
\begin{tabular}{>{\hspace{0pt}}m{0.020\linewidth}>{\hspace{0pt}}m{0.045\linewidth}|>{\centering\hspace{0pt}}m{0.069\linewidth}>{\centering\hspace{0pt}}m{0.069\linewidth}>{\centering\hspace{0pt}}m{0.069\linewidth}>{\centering\hspace{0pt}}m{0.069\linewidth}>{\centering\hspace{0pt}}m{0.069\linewidth}>{\centering\hspace{0pt}}m{0.069\linewidth}|>{\centering\hspace{0pt}}m{0.058\linewidth}>{\centering\hspace{0pt}}m{0.058\linewidth}>{\centering\hspace{0pt}}m{0.058\linewidth}>{\centering\hspace{0pt}}m{0.069\linewidth}>{\centering\hspace{0pt}}m{0.069\linewidth}>{\centering\arraybackslash\hspace{0pt}}m{0.069\linewidth}} 
\hline
 &  & \multicolumn{6}{>{\centering\hspace{0pt}}m{0.39\linewidth}|}{$G_S$  ($x'$) $\rightarrow$} & \multicolumn{6}{>{\centering\arraybackslash\hspace{0pt}}m{0.39\linewidth}}{$S_S$ ($\hat{x}$) $\rightarrow$} \\ 
\hline\multicolumn{1}{>{\centering\hspace{0pt}}m{0.05\linewidth}}{\textbf{Set}} & \textbf{$x$} $\downarrow$ & \textbf{en} & \textbf{fr} & \textbf{es} & \textbf{hi} & \textbf{gu} & \textbf{bn} & \textbf{en} & \textbf{fr} & \textbf{es} & \textbf{hi} & \textbf{gu} & \textbf{bn} \\ \hline\multicolumn{1}{>{\centering\hspace{0pt}}m{0.05\linewidth}}{\textbf{IL}}& en & \cellcolor[HTML]{78B2E4}\textbf{76.19}& \cellcolor[HTML]{92C0E9}69.32& \cellcolor[HTML]{8CBDE8}70.75& \cellcolor[HTML]{D0E4F5}52.47& \cellcolor[HTML]{E1EEF9}47.95& \cellcolor[HTML]{D7E8F7}50.54& \cellcolor[HTML]{4C99DB}94.07& \cellcolor[HTML]{3F91D9}\textbf{95.77}& \cellcolor[HTML]{4595DA}94.91& \cellcolor[HTML]{61A5DF}91.2& \cellcolor[HTML]{3F92D9}95.73& \cellcolor[HTML]{60A4DF}91.34 \\ 
& fr & \cellcolor[HTML]{89BBE7}71.67& \cellcolor[HTML]{6CABE2}\textbf{79.3}& \cellcolor[HTML]{7EB5E5}74.6& \cellcolor[HTML]{D5E7F6}51.05& \cellcolor[HTML]{E0EDF8}48.2& \cellcolor[HTML]{DAEAF7}49.87& \cellcolor[HTML]{4293D9}95.37& \cellcolor[HTML]{4896DA}94.61& \cellcolor[HTML]{4B98DB}94.17& \cellcolor[HTML]{59A0DE}92.22& \cellcolor[HTML]{3B8FD8}\textbf{96.33}& \cellcolor[HTML]{579FDE}92.5 \\
\hline 

\end{tabular}
}
\caption{\label{tab:mend_bloom-560M_mixed}
The table represents the $G_S$ and $S_S$ using \textbf{MEND} over fine-tuned \texttt{BLOOM} on the fever `\textbf{mixed}' dataset. 
}
\end{table*} 
%%%%%%%%%%%%%%%%%%%%%%%% MEND-BLOOM-inverse %%%%%%%%%%%%%%%%%%%%%%%%%%%% 
\begin{table*}[htpb]
\centering

\resizebox{\linewidth}{!}{%
\begin{tabular}{>{\hspace{0pt}}m{0.020\linewidth}>{\hspace{0pt}}m{0.045\linewidth}|>{\centering\hspace{0pt}}m{0.069\linewidth}>{\centering\hspace{0pt}}m{0.069\linewidth}>{\centering\hspace{0pt}}m{0.069\linewidth}>{\centering\hspace{0pt}}m{0.069\linewidth}>{\centering\hspace{0pt}}m{0.069\linewidth}>{\centering\hspace{0pt}}m{0.069\linewidth}|>{\centering\hspace{0pt}}m{0.058\linewidth}>{\centering\hspace{0pt}}m{0.058\linewidth}>{\centering\hspace{0pt}}m{0.058\linewidth}>{\centering\hspace{0pt}}m{0.069\linewidth}>{\centering\hspace{0pt}}m{0.069\linewidth}>{\centering\arraybackslash\hspace{0pt}}m{0.069\linewidth}} 
\hline
 &  & \multicolumn{6}{>{\centering\hspace{0pt}}m{0.39\linewidth}|}{$G_S$  ($x'$) $\rightarrow$} & \multicolumn{6}{>{\centering\arraybackslash\hspace{0pt}}m{0.39\linewidth}}{$S_S$ ($\hat{x}$) $\rightarrow$} \\ 
\hline\multicolumn{1}{>{\centering\hspace{0pt}}m{0.05\linewidth}}{\textbf{Set}} & \textbf{$x$} $\downarrow$ & \textbf{en} & \textbf{fr} & \textbf{es} & \textbf{hi} & \textbf{gu} & \textbf{bn} & \textbf{en} & \textbf{fr} & \textbf{es} & \textbf{hi} & \textbf{gu} & \textbf{bn} \\ \hline\multicolumn{1}{>{\centering\hspace{0pt}}m{0.05\linewidth}}{\textbf{IL}}& en & \cellcolor[HTML]{67A8E1}\textbf{80.81}& \cellcolor[HTML]{7AB3E4}75.78& \cellcolor[HTML]{85BAE7}72.59& \cellcolor[HTML]{92C1E9}69.15& \cellcolor[HTML]{85B9E6}72.76& \cellcolor[HTML]{91C0E9}69.57& \cellcolor[HTML]{8FBFE8}85.06& \cellcolor[HTML]{87BBE7}86.04& \cellcolor[HTML]{79B2E4}87.97& \cellcolor[HTML]{5CA2DE}\textbf{91.87}& \cellcolor[HTML]{74B0E3}88.62& \cellcolor[HTML]{5EA3DF}91.64 \\ 
& fr & \cellcolor[HTML]{72AFE3}77.79& \cellcolor[HTML]{62A5E0}\textbf{82.06}& \cellcolor[HTML]{83B8E6}73.18& \cellcolor[HTML]{8DBEE8}70.41& \cellcolor[HTML]{84B9E6}72.92& \cellcolor[HTML]{89BCE7}71.5& \cellcolor[HTML]{90BFE9}84.89& \cellcolor[HTML]{8BBCE8}85.6& \cellcolor[HTML]{7AB3E4}87.8& \cellcolor[HTML]{5DA2DF}\textbf{91.74}& \cellcolor[HTML]{7FB6E5}87.13& \cellcolor[HTML]{61A5E0}91.14 \\
\hline 

\end{tabular}
}
\caption{\label{tab:mend_bloom-560M_inverse}
The table represents the $G_S$ and $S_S$ using \textbf{MEND} over fine-tuned \texttt{BLOOM} on the fever `\textbf{inverse}' dataset. 
}
\end{table*} 
%%%%%%%%%%%%%%%%%%%%%%%% TABLE START MEND-mBERT %%%%%%%%%%%%%%%%%%%%%%%%%%%% 
%%%%%%%%%%%%%%%%%%%%%%%% MEND-mBERT-en %%%%%%%%%%%%%%%%%%%%%%%%%%%% 
\begin{table*}[htpb]
\centering

\resizebox{\linewidth}{!}{%
\begin{tabular}{>{\hspace{0pt}}m{0.020\linewidth}>{\hspace{0pt}}m{0.045\linewidth}|>{\centering\hspace{0pt}}m{0.069\linewidth}>{\centering\hspace{0pt}}m{0.069\linewidth}>{\centering\hspace{0pt}}m{0.069\linewidth}>{\centering\hspace{0pt}}m{0.069\linewidth}>{\centering\hspace{0pt}}m{0.069\linewidth}>{\centering\hspace{0pt}}m{0.069\linewidth}|>{\centering\hspace{0pt}}m{0.058\linewidth}>{\centering\hspace{0pt}}m{0.058\linewidth}>{\centering\hspace{0pt}}m{0.058\linewidth}>{\centering\hspace{0pt}}m{0.069\linewidth}>{\centering\hspace{0pt}}m{0.069\linewidth}>{\centering\arraybackslash\hspace{0pt}}m{0.069\linewidth}} 
\hline
 &  & \multicolumn{6}{>{\centering\hspace{0pt}}m{0.39\linewidth}|}{$G_S$  ($x'$) $\rightarrow$} & \multicolumn{6}{>{\centering\arraybackslash\hspace{0pt}}m{0.39\linewidth}}{$S_S$ ($\hat{x}$) $\rightarrow$} \\ 
\hline\multicolumn{1}{>{\centering\hspace{0pt}}m{0.05\linewidth}}{\textbf{Set}} & \textbf{$x$} $\downarrow$ & \textbf{en} & \textbf{fr} & \textbf{es} & \textbf{hi} & \textbf{gu} & \textbf{bn} & \textbf{en} & \textbf{fr} & \textbf{es} & \textbf{hi} & \textbf{gu} & \textbf{bn} \\ \hline\multicolumn{1}{>{\centering\hspace{0pt}}m{0.05\linewidth}}{\textbf{IL}}& en & \cellcolor[HTML]{3E91D9}\textbf{91.79}& \cellcolor[HTML]{4E9ADC}87.51& \cellcolor[HTML]{4D99DB}87.85& \cellcolor[HTML]{B8D6F1}58.93& \cellcolor[HTML]{D0E4F5}52.56& \cellcolor[HTML]{C6DEF3}55.24& \cellcolor[HTML]{2C87D5}98.32& \cellcolor[HTML]{2E88D5}98.09& \cellcolor[HTML]{2B86D5}\textbf{98.41}& \cellcolor[HTML]{3089D6}97.76& \cellcolor[HTML]{2D87D5}98.2& \cellcolor[HTML]{328AD6}97.48 \\ 
& fr & \cellcolor[HTML]{4193D9}90.86& \cellcolor[HTML]{2B86D5}\textbf{96.9}& \cellcolor[HTML]{3B8FD8}92.54& \cellcolor[HTML]{B9D7F1}58.59& \cellcolor[HTML]{D2E5F6}51.89& \cellcolor[HTML]{C4DDF3}55.83& \cellcolor[HTML]{2985D4}\textbf{98.76}& \cellcolor[HTML]{3089D6}97.72& \cellcolor[HTML]{2B86D5}98.43& \cellcolor[HTML]{2C87D5}98.26& \cellcolor[HTML]{2B86D5}98.45& \cellcolor[HTML]{2F88D6}97.92 \\
\hline 

\end{tabular}
}
\caption{\label{tab:mend_mBERT_en}
The table represents the $G_S$ and $S_S$ using \textbf{MEND} over fine-tuned \texttt{mBERT} on the fever `\textbf{en}' dataset. 
}
\end{table*} 
%%%%%%%%%%%%%%%%%%%%%%%% MEND-mBERT-fr %%%%%%%%%%%%%%%%%%%%%%%%%%%% 
\begin{table*}[htpb]
\centering

\resizebox{\linewidth}{!}{%
\begin{tabular}{>{\hspace{0pt}}m{0.020\linewidth}>{\hspace{0pt}}m{0.045\linewidth}|>{\centering\hspace{0pt}}m{0.069\linewidth}>{\centering\hspace{0pt}}m{0.069\linewidth}>{\centering\hspace{0pt}}m{0.069\linewidth}>{\centering\hspace{0pt}}m{0.069\linewidth}>{\centering\hspace{0pt}}m{0.069\linewidth}>{\centering\hspace{0pt}}m{0.069\linewidth}|>{\centering\hspace{0pt}}m{0.058\linewidth}>{\centering\hspace{0pt}}m{0.058\linewidth}>{\centering\hspace{0pt}}m{0.058\linewidth}>{\centering\hspace{0pt}}m{0.069\linewidth}>{\centering\hspace{0pt}}m{0.069\linewidth}>{\centering\arraybackslash\hspace{0pt}}m{0.069\linewidth}} 
\hline
 &  & \multicolumn{6}{>{\centering\hspace{0pt}}m{0.39\linewidth}|}{$G_S$  ($x'$) $\rightarrow$} & \multicolumn{6}{>{\centering\arraybackslash\hspace{0pt}}m{0.39\linewidth}}{$S_S$ ($\hat{x}$) $\rightarrow$} \\ 
\hline\multicolumn{1}{>{\centering\hspace{0pt}}m{0.05\linewidth}}{\textbf{Set}} & \textbf{$x$} $\downarrow$ & \textbf{en} & \textbf{fr} & \textbf{es} & \textbf{hi} & \textbf{gu} & \textbf{bn} & \textbf{en} & \textbf{fr} & \textbf{es} & \textbf{hi} & \textbf{gu} & \textbf{bn} \\ \hline\multicolumn{1}{>{\centering\hspace{0pt}}m{0.05\linewidth}}{\textbf{IL}}& en & \cellcolor[HTML]{2080D3}\textbf{99.41}& \cellcolor[HTML]{2F89D6}95.73& \cellcolor[HTML]{348BD7}94.47& \cellcolor[HTML]{C7DFF4}54.82& \cellcolor[HTML]{D1E5F6}52.14& \cellcolor[HTML]{D1E5F6}52.22& \cellcolor[HTML]{2080D3}99.35& \cellcolor[HTML]{2080D3}99.77& \cellcolor[HTML]{2080D3}99.79& \cellcolor[HTML]{2080D3}99.92& \cellcolor[HTML]{2080D3}\textbf{99.96}& \cellcolor[HTML]{2080D3}99.92 \\ 
& fr & \cellcolor[HTML]{3089D6}95.64& \cellcolor[HTML]{2080D3}\textbf{99.5}& \cellcolor[HTML]{328AD6}94.97& \cellcolor[HTML]{CAE0F4}54.23& \cellcolor[HTML]{D2E5F6}52.05& \cellcolor[HTML]{D2E5F6}52.05& \cellcolor[HTML]{2080D3}99.85& \cellcolor[HTML]{2080D3}99.62& \cellcolor[HTML]{2080D3}99.87& \cellcolor[HTML]{2080D3}99.98& \cellcolor[HTML]{2080D3}99.98& \cellcolor[HTML]{2080D3}\textbf{100.0} \\
\hline 

\end{tabular}
}
\caption{\label{tab:mend_mBERT_fr}
The table represents the $G_S$ and $S_S$ using \textbf{MEND} over fine-tuned \texttt{mBERT} on the fever `\textbf{fr}' dataset. 
}
\end{table*} 
%%%%%%%%%%%%%%%%%%%%%%%% MEND-mBERT-es %%%%%%%%%%%%%%%%%%%%%%%%%%%% 
\begin{table*}[htpb]
\centering

\resizebox{\linewidth}{!}{%
\begin{tabular}{>{\hspace{0pt}}m{0.020\linewidth}>{\hspace{0pt}}m{0.045\linewidth}|>{\centering\hspace{0pt}}m{0.069\linewidth}>{\centering\hspace{0pt}}m{0.069\linewidth}>{\centering\hspace{0pt}}m{0.069\linewidth}>{\centering\hspace{0pt}}m{0.069\linewidth}>{\centering\hspace{0pt}}m{0.069\linewidth}>{\centering\hspace{0pt}}m{0.069\linewidth}|>{\centering\hspace{0pt}}m{0.058\linewidth}>{\centering\hspace{0pt}}m{0.058\linewidth}>{\centering\hspace{0pt}}m{0.058\linewidth}>{\centering\hspace{0pt}}m{0.069\linewidth}>{\centering\hspace{0pt}}m{0.069\linewidth}>{\centering\arraybackslash\hspace{0pt}}m{0.069\linewidth}} 
\hline
 &  & \multicolumn{6}{>{\centering\hspace{0pt}}m{0.39\linewidth}|}{$G_S$  ($x'$) $\rightarrow$} & \multicolumn{6}{>{\centering\arraybackslash\hspace{0pt}}m{0.39\linewidth}}{$S_S$ ($\hat{x}$) $\rightarrow$} \\ 
\hline\multicolumn{1}{>{\centering\hspace{0pt}}m{0.05\linewidth}}{\textbf{Set}} & \textbf{$x$} $\downarrow$ & \textbf{en} & \textbf{fr} & \textbf{es} & \textbf{hi} & \textbf{gu} & \textbf{bn} & \textbf{en} & \textbf{fr} & \textbf{es} & \textbf{hi} & \textbf{gu} & \textbf{bn} \\ \hline\multicolumn{1}{>{\centering\hspace{0pt}}m{0.05\linewidth}}{\textbf{IL}}& en & \cellcolor[HTML]{2080D3}\textbf{100.0}& \cellcolor[HTML]{2583D4}98.49& \cellcolor[HTML]{2E88D5}96.06& \cellcolor[HTML]{C5DEF3}55.57& \cellcolor[HTML]{D0E4F5}52.56& \cellcolor[HTML]{CEE3F5}52.98& \cellcolor[HTML]{2C86D5}98.37& \cellcolor[HTML]{2080D3}99.35& \cellcolor[HTML]{2080D3}99.73& \cellcolor[HTML]{2080D3}\textbf{99.98}& \cellcolor[HTML]{2080D3}\textbf{99.98}& \cellcolor[HTML]{2080D3}\textbf{99.98} \\ 
& fr & \cellcolor[HTML]{2A85D5}97.23& \cellcolor[HTML]{2080D3}\textbf{99.92}& \cellcolor[HTML]{2D87D5}96.4& \cellcolor[HTML]{C5DEF3}55.49& \cellcolor[HTML]{D1E5F6}52.22& \cellcolor[HTML]{D0E4F5}52.47& \cellcolor[HTML]{2080D3}99.77& \cellcolor[HTML]{4796DA}94.7& \cellcolor[HTML]{2080D3}99.79& \cellcolor[HTML]{2080D3}99.96& \cellcolor[HTML]{2080D3}\textbf{100.0}& \cellcolor[HTML]{2080D3}\textbf{100.0} \\
\hline 

\end{tabular}
}
\caption{\label{tab:mend_mBERT_es}
The table represents the $G_S$ and $S_S$ using \textbf{MEND} over fine-tuned \texttt{mBERT} on the fever `\textbf{es}' dataset. 
}
\end{table*} 
%%%%%%%%%%%%%%%%%%%%%%%% MEND-mBERT-hi %%%%%%%%%%%%%%%%%%%%%%%%%%%% 
\begin{table*}[htpb]
\centering

\resizebox{\linewidth}{!}{%
\begin{tabular}{>{\hspace{0pt}}m{0.020\linewidth}>{\hspace{0pt}}m{0.045\linewidth}|>{\centering\hspace{0pt}}m{0.069\linewidth}>{\centering\hspace{0pt}}m{0.069\linewidth}>{\centering\hspace{0pt}}m{0.069\linewidth}>{\centering\hspace{0pt}}m{0.069\linewidth}>{\centering\hspace{0pt}}m{0.069\linewidth}>{\centering\hspace{0pt}}m{0.069\linewidth}|>{\centering\hspace{0pt}}m{0.058\linewidth}>{\centering\hspace{0pt}}m{0.058\linewidth}>{\centering\hspace{0pt}}m{0.058\linewidth}>{\centering\hspace{0pt}}m{0.069\linewidth}>{\centering\hspace{0pt}}m{0.069\linewidth}>{\centering\arraybackslash\hspace{0pt}}m{0.069\linewidth}} 
\hline
 &  & \multicolumn{6}{>{\centering\hspace{0pt}}m{0.39\linewidth}|}{$G_S$  ($x'$) $\rightarrow$} & \multicolumn{6}{>{\centering\arraybackslash\hspace{0pt}}m{0.39\linewidth}}{$S_S$ ($\hat{x}$) $\rightarrow$} \\ 
\hline\multicolumn{1}{>{\centering\hspace{0pt}}m{0.05\linewidth}}{\textbf{Set}} & \textbf{$x$} $\downarrow$ & \textbf{en} & \textbf{fr} & \textbf{es} & \textbf{hi} & \textbf{gu} & \textbf{bn} & \textbf{en} & \textbf{fr} & \textbf{es} & \textbf{hi} & \textbf{gu} & \textbf{bn} \\ \hline\multicolumn{1}{>{\centering\hspace{0pt}}m{0.05\linewidth}}{\textbf{IL}}& en & \cellcolor[HTML]{4595DA}\textbf{90.03}& \cellcolor[HTML]{4C99DB}87.93& \cellcolor[HTML]{4F9ADC}87.26& \cellcolor[HTML]{C2DCF3}56.16& \cellcolor[HTML]{CCE2F5}53.48& \cellcolor[HTML]{CBE1F4}53.73& \cellcolor[HTML]{A1C9EC}82.52& \cellcolor[HTML]{529CDC}93.23& \cellcolor[HTML]{60A4DF}91.37& \cellcolor[HTML]{2080D3}99.06& \cellcolor[HTML]{2080D3}99.08& \cellcolor[HTML]{2080D3}\textbf{99.1} \\ 
& fr & \cellcolor[HTML]{378DD7}93.55& \cellcolor[HTML]{3089D6}\textbf{95.64}& \cellcolor[HTML]{338BD6}94.8& \cellcolor[HTML]{C6DEF3}55.32& \cellcolor[HTML]{D0E4F5}52.56& \cellcolor[HTML]{CFE4F5}52.72& \cellcolor[HTML]{82B7E6}86.8& \cellcolor[HTML]{83B8E6}86.61& \cellcolor[HTML]{579FDD}92.52& \cellcolor[HTML]{2080D3}99.62& \cellcolor[HTML]{2080D3}99.64& \cellcolor[HTML]{2080D3}\textbf{99.73} \\
\hline 

\end{tabular}
}
\caption{\label{tab:mend_mBERT_hi}
The table represents the $G_S$ and $S_S$ using \textbf{MEND} over fine-tuned \texttt{mBERT} on the fever `\textbf{hi}' dataset. 
}
\end{table*} 
%%%%%%%%%%%%%%%%%%%%%%%% MEND-mBERT-gu %%%%%%%%%%%%%%%%%%%%%%%%%%%% 
\begin{table*}[htpb]
\centering

\resizebox{\linewidth}{!}{%
\begin{tabular}{>{\hspace{0pt}}m{0.020\linewidth}>{\hspace{0pt}}m{0.045\linewidth}|>{\centering\hspace{0pt}}m{0.069\linewidth}>{\centering\hspace{0pt}}m{0.069\linewidth}>{\centering\hspace{0pt}}m{0.069\linewidth}>{\centering\hspace{0pt}}m{0.069\linewidth}>{\centering\hspace{0pt}}m{0.069\linewidth}>{\centering\hspace{0pt}}m{0.069\linewidth}|>{\centering\hspace{0pt}}m{0.058\linewidth}>{\centering\hspace{0pt}}m{0.058\linewidth}>{\centering\hspace{0pt}}m{0.058\linewidth}>{\centering\hspace{0pt}}m{0.069\linewidth}>{\centering\hspace{0pt}}m{0.069\linewidth}>{\centering\arraybackslash\hspace{0pt}}m{0.069\linewidth}} 
\hline
 &  & \multicolumn{6}{>{\centering\hspace{0pt}}m{0.39\linewidth}|}{$G_S$  ($x'$) $\rightarrow$} & \multicolumn{6}{>{\centering\arraybackslash\hspace{0pt}}m{0.39\linewidth}}{$S_S$ ($\hat{x}$) $\rightarrow$} \\ 
\hline\multicolumn{1}{>{\centering\hspace{0pt}}m{0.05\linewidth}}{\textbf{Set}} & \textbf{$x$} $\downarrow$ & \textbf{en} & \textbf{fr} & \textbf{es} & \textbf{hi} & \textbf{gu} & \textbf{bn} & \textbf{en} & \textbf{fr} & \textbf{es} & \textbf{hi} & \textbf{gu} & \textbf{bn} \\ \hline\multicolumn{1}{>{\centering\hspace{0pt}}m{0.05\linewidth}}{\textbf{IL}}& en & \cellcolor[HTML]{5BA2DE}\textbf{83.91}& \cellcolor[HTML]{69A9E1}80.3& \cellcolor[HTML]{68A9E1}80.47& \cellcolor[HTML]{BAD8F1}58.42& \cellcolor[HTML]{CFE3F5}52.89& \cellcolor[HTML]{C9E0F4}54.48& \cellcolor[HTML]{9BC6EB}83.32& \cellcolor[HTML]{96C3EA}84.05& \cellcolor[HTML]{84B9E6}86.48& \cellcolor[HTML]{2985D4}98.78& \cellcolor[HTML]{2080D3}\textbf{99.35}& \cellcolor[HTML]{2080D3}99.08 \\ 
& fr & \cellcolor[HTML]{75B0E3}76.95& \cellcolor[HTML]{64A6E0}\textbf{81.64}& \cellcolor[HTML]{6FADE2}78.71& \cellcolor[HTML]{BDD9F2}57.5& \cellcolor[HTML]{CEE3F5}53.06& \cellcolor[HTML]{CBE1F4}53.9& \cellcolor[HTML]{6FADE2}89.25& \cellcolor[HTML]{A5CCED}81.98& \cellcolor[HTML]{80B6E5}87.05& \cellcolor[HTML]{2985D4}98.78& \cellcolor[HTML]{2080D3}\textbf{99.33}& \cellcolor[HTML]{2080D3}99.18 \\
\hline 

\end{tabular}
}
\caption{\label{tab:mend_mBERT_gu}
The table represents the $G_S$ and $S_S$ using \textbf{MEND} over fine-tuned \texttt{mBERT} on the fever `\textbf{gu}' dataset. 
}
\end{table*} 
%%%%%%%%%%%%%%%%%%%%%%%% MEND-mBERT-bn %%%%%%%%%%%%%%%%%%%%%%%%%%%% 
\begin{table*}[htpb]
\centering

\resizebox{\linewidth}{!}{%
\begin{tabular}{>{\hspace{0pt}}m{0.020\linewidth}>{\hspace{0pt}}m{0.045\linewidth}|>{\centering\hspace{0pt}}m{0.069\linewidth}>{\centering\hspace{0pt}}m{0.069\linewidth}>{\centering\hspace{0pt}}m{0.069\linewidth}>{\centering\hspace{0pt}}m{0.069\linewidth}>{\centering\hspace{0pt}}m{0.069\linewidth}>{\centering\hspace{0pt}}m{0.069\linewidth}|>{\centering\hspace{0pt}}m{0.058\linewidth}>{\centering\hspace{0pt}}m{0.058\linewidth}>{\centering\hspace{0pt}}m{0.058\linewidth}>{\centering\hspace{0pt}}m{0.069\linewidth}>{\centering\hspace{0pt}}m{0.069\linewidth}>{\centering\arraybackslash\hspace{0pt}}m{0.069\linewidth}} 
\hline
 &  & \multicolumn{6}{>{\centering\hspace{0pt}}m{0.39\linewidth}|}{$G_S$  ($x'$) $\rightarrow$} & \multicolumn{6}{>{\centering\arraybackslash\hspace{0pt}}m{0.39\linewidth}}{$S_S$ ($\hat{x}$) $\rightarrow$} \\ 
\hline\multicolumn{1}{>{\centering\hspace{0pt}}m{0.05\linewidth}}{\textbf{Set}} & \textbf{$x$} $\downarrow$ & \textbf{en} & \textbf{fr} & \textbf{es} & \textbf{hi} & \textbf{gu} & \textbf{bn} & \textbf{en} & \textbf{fr} & \textbf{es} & \textbf{hi} & \textbf{gu} & \textbf{bn} \\ \hline\multicolumn{1}{>{\centering\hspace{0pt}}m{0.05\linewidth}}{\textbf{IL}}& en & \cellcolor[HTML]{2080D3}\textbf{99.25}& \cellcolor[HTML]{2B86D5}96.98& \cellcolor[HTML]{2683D4}98.24& \cellcolor[HTML]{C2DCF3}56.16& \cellcolor[HTML]{CFE4F5}52.72& \cellcolor[HTML]{D0E4F5}52.64& \cellcolor[HTML]{77B1E4}88.24& \cellcolor[HTML]{5CA2DE}91.83& \cellcolor[HTML]{6AAAE1}90.03& \cellcolor[HTML]{2080D3}99.56& \cellcolor[HTML]{2080D3}\textbf{99.67}& \cellcolor[HTML]{2080D3}99.56 \\ 
& fr & \cellcolor[HTML]{368CD7}93.88& \cellcolor[HTML]{2A85D5}\textbf{97.23}& \cellcolor[HTML]{318AD6}95.22& \cellcolor[HTML]{C7DFF4}54.99& \cellcolor[HTML]{D0E4F5}52.64& \cellcolor[HTML]{D0E4F5}52.64& \cellcolor[HTML]{4494DA}95.12& \cellcolor[HTML]{73AFE3}88.81& \cellcolor[HTML]{5DA3DF}91.7& \cellcolor[HTML]{2080D3}99.56& \cellcolor[HTML]{2080D3}\textbf{99.62}& \cellcolor[HTML]{2080D3}99.54 \\
\hline 

\end{tabular}
}
\caption{\label{tab:mend_mBERT_bn}
The table represents the $G_S$ and $S_S$ using \textbf{MEND} over fine-tuned \texttt{mBERT} on the fever `\textbf{bn}' dataset. 
}
\end{table*} 
%%%%%%%%%%%%%%%%%%%%%%%% MEND-mBERT-mixed %%%%%%%%%%%%%%%%%%%%%%%%%%%% 
\begin{table*}[htpb]
\centering

\resizebox{\linewidth}{!}{%
\begin{tabular}{>{\hspace{0pt}}m{0.020\linewidth}>{\hspace{0pt}}m{0.045\linewidth}|>{\centering\hspace{0pt}}m{0.069\linewidth}>{\centering\hspace{0pt}}m{0.069\linewidth}>{\centering\hspace{0pt}}m{0.069\linewidth}>{\centering\hspace{0pt}}m{0.069\linewidth}>{\centering\hspace{0pt}}m{0.069\linewidth}>{\centering\hspace{0pt}}m{0.069\linewidth}|>{\centering\hspace{0pt}}m{0.058\linewidth}>{\centering\hspace{0pt}}m{0.058\linewidth}>{\centering\hspace{0pt}}m{0.058\linewidth}>{\centering\hspace{0pt}}m{0.069\linewidth}>{\centering\hspace{0pt}}m{0.069\linewidth}>{\centering\arraybackslash\hspace{0pt}}m{0.069\linewidth}} 
\hline
 &  & \multicolumn{6}{>{\centering\hspace{0pt}}m{0.39\linewidth}|}{$G_S$  ($x'$) $\rightarrow$} & \multicolumn{6}{>{\centering\arraybackslash\hspace{0pt}}m{0.39\linewidth}}{$S_S$ ($\hat{x}$) $\rightarrow$} \\ 
\hline\multicolumn{1}{>{\centering\hspace{0pt}}m{0.05\linewidth}}{\textbf{Set}} & \textbf{$x$} $\downarrow$ & \textbf{en} & \textbf{fr} & \textbf{es} & \textbf{hi} & \textbf{gu} & \textbf{bn} & \textbf{en} & \textbf{fr} & \textbf{es} & \textbf{hi} & \textbf{gu} & \textbf{bn} \\ \hline\multicolumn{1}{>{\centering\hspace{0pt}}m{0.05\linewidth}}{\textbf{IL}}& en & \cellcolor[HTML]{4192D9}\textbf{91.03}& \cellcolor[HTML]{539DDD}86.25& \cellcolor[HTML]{539DDD}86.25& \cellcolor[HTML]{AFD1EF}61.27& \cellcolor[HTML]{CEE3F5}52.98& \cellcolor[HTML]{BEDAF2}57.42& \cellcolor[HTML]{328AD6}97.46& \cellcolor[HTML]{348BD7}97.23& \cellcolor[HTML]{318AD6}97.59& \cellcolor[HTML]{2F88D6}97.92& \cellcolor[HTML]{2985D4}\textbf{98.74}& \cellcolor[HTML]{3089D6}97.84 \\ 
& fr & \cellcolor[HTML]{3D90D8}92.12& \cellcolor[HTML]{2884D4}\textbf{97.65}& \cellcolor[HTML]{358CD7}94.3& \cellcolor[HTML]{B0D2EF}61.11& \cellcolor[HTML]{CFE4F5}52.72& \cellcolor[HTML]{C0DBF2}56.83& \cellcolor[HTML]{2E87D5}98.11& \cellcolor[HTML]{388DD7}96.75& \cellcolor[HTML]{2D87D5}98.13& \cellcolor[HTML]{2E88D5}98.03& \cellcolor[HTML]{2884D4}\textbf{98.87}& \cellcolor[HTML]{2F88D5}97.97 \\
\hline 

\end{tabular}
}
\caption{\label{tab:mend_mBERT_mixed}
The table represents the $G_S$ and $S_S$ using \textbf{MEND} over fine-tuned \texttt{mBERT} on the fever `\textbf{mixed}' dataset. 
}
\end{table*}

\include{supplementary2}

%%%%%%%%%%%%%%%%%%%%%%%% FT-XLM-RoBERTa-gu %%%%%%%%%%%%%%%%%%%%%%%%%%%% 
\begin{table*}[htpb]
\centering

\resizebox{\linewidth}{!}{%
\begin{tabular}{>{\hspace{0pt}}m{0.020\linewidth}>{\hspace{0pt}}m{0.045\linewidth}|>{\centering\hspace{0pt}}m{0.069\linewidth}>{\centering\hspace{0pt}}m{0.069\linewidth}>{\centering\hspace{0pt}}m{0.069\linewidth}>{\centering\hspace{0pt}}m{0.069\linewidth}>{\centering\hspace{0pt}}m{0.069\linewidth}>{\centering\hspace{0pt}}m{0.069\linewidth}|>{\centering\hspace{0pt}}m{0.058\linewidth}>{\centering\hspace{0pt}}m{0.058\linewidth}>{\centering\hspace{0pt}}m{0.058\linewidth}>{\centering\hspace{0pt}}m{0.069\linewidth}>{\centering\hspace{0pt}}m{0.069\linewidth}>{\centering\arraybackslash\hspace{0pt}}m{0.069\linewidth}} 
\hline
 &  & \multicolumn{6}{>{\centering\hspace{0pt}}m{0.39\linewidth}|}{$G_S$  ($x'$) $\rightarrow$} & \multicolumn{6}{>{\centering\arraybackslash\hspace{0pt}}m{0.39\linewidth}}{$S_S$ ($\hat{x}$) $\rightarrow$} \\ 
\hline\multicolumn{1}{>{\centering\hspace{0pt}}m{0.05\linewidth}}{\textbf{Set}} & \textbf{$x$} $\downarrow$ & \textbf{en} & \textbf{fr} & \textbf{es} & \textbf{hi} & \textbf{gu} & \textbf{bn} & \textbf{en} & \textbf{fr} & \textbf{es} & \textbf{hi} & \textbf{gu} & \textbf{bn} \\ \hline\multicolumn{1}{>{\centering\hspace{0pt}}m{0.05\linewidth}}{\textbf{IL}}& en & \cellcolor[HTML]{4C99DB}\textbf{88.01}& \cellcolor[HTML]{7DB5E5}74.85& \cellcolor[HTML]{76B1E4}76.61& \cellcolor[HTML]{C5DEF3}55.41& \cellcolor[HTML]{D1E5F6}52.14& \cellcolor[HTML]{CFE3F5}52.89& \cellcolor[HTML]{92C1E9}84.62& \cellcolor[HTML]{70AEE2}89.12& \cellcolor[HTML]{83B8E6}86.67& \cellcolor[HTML]{338AD6}97.4& \cellcolor[HTML]{2080D3}\textbf{99.96}& \cellcolor[HTML]{2885D4}98.81 \\ 
& fr & \cellcolor[HTML]{65A7E0}81.31& \cellcolor[HTML]{4293D9}\textbf{90.7}& \cellcolor[HTML]{64A7E0}81.47& \cellcolor[HTML]{C8E0F4}54.57& \cellcolor[HTML]{D2E5F6}52.05& \cellcolor[HTML]{D0E4F5}52.56& \cellcolor[HTML]{8BBDE8}85.52& \cellcolor[HTML]{7AB3E4}87.87& \cellcolor[HTML]{86BAE7}86.23& \cellcolor[HTML]{2E88D5}97.99& \cellcolor[HTML]{2080D3}\textbf{99.98}& \cellcolor[HTML]{2080D3}99.12 \\
\hline 

\end{tabular}
}
\caption{\label{tab:ft_XLM-RoBERTa_gu}
The table represents the $G_S$ and $S_S$ using \textbf{FT} over fine-tuned \texttt{XLM-RoBERTa} on the fever `\textbf{gu}' dataset. 
}
\end{table*}
%%%%%%%%%%%%%%%%%%%%%%%% FT-XLM-RoBERTa-bn %%%%%%%%%%%%%%%%%%%%%%%%%%%% 
\begin{table*}[htpb]
\centering

\resizebox{\linewidth}{!}{%
\begin{tabular}{>{\hspace{0pt}}m{0.020\linewidth}>{\hspace{0pt}}m{0.045\linewidth}|>{\centering\hspace{0pt}}m{0.069\linewidth}>{\centering\hspace{0pt}}m{0.069\linewidth}>{\centering\hspace{0pt}}m{0.069\linewidth}>{\centering\hspace{0pt}}m{0.069\linewidth}>{\centering\hspace{0pt}}m{0.069\linewidth}>{\centering\hspace{0pt}}m{0.069\linewidth}|>{\centering\hspace{0pt}}m{0.058\linewidth}>{\centering\hspace{0pt}}m{0.058\linewidth}>{\centering\hspace{0pt}}m{0.058\linewidth}>{\centering\hspace{0pt}}m{0.069\linewidth}>{\centering\hspace{0pt}}m{0.069\linewidth}>{\centering\arraybackslash\hspace{0pt}}m{0.069\linewidth}} 
\hline
 &  & \multicolumn{6}{>{\centering\hspace{0pt}}m{0.39\linewidth}|}{$G_S$  ($x'$) $\rightarrow$} & \multicolumn{6}{>{\centering\arraybackslash\hspace{0pt}}m{0.39\linewidth}}{$S_S$ ($\hat{x}$) $\rightarrow$} \\ 
\hline\multicolumn{1}{>{\centering\hspace{0pt}}m{0.05\linewidth}}{\textbf{Set}} & \textbf{$x$} $\downarrow$ & \textbf{en} & \textbf{fr} & \textbf{es} & \textbf{hi} & \textbf{gu} & \textbf{bn} & \textbf{en} & \textbf{fr} & \textbf{es} & \textbf{hi} & \textbf{gu} & \textbf{bn} \\ \hline\multicolumn{1}{>{\centering\hspace{0pt}}m{0.05\linewidth}}{\textbf{IL}}& en & \cellcolor[HTML]{D0E4F5}\textbf{52.39}& \cellcolor[HTML]{D2E5F6}52.05& \cellcolor[HTML]{D2E5F6}52.05& \cellcolor[HTML]{D2E5F6}52.05& \cellcolor[HTML]{D2E5F6}52.05& \cellcolor[HTML]{D2E5F6}52.05& \cellcolor[HTML]{2080D3}\textbf{100.0}& \cellcolor[HTML]{2080D3}\textbf{100.0}& \cellcolor[HTML]{2080D3}\textbf{100.0}& \cellcolor[HTML]{2080D3}\textbf{100.0}& \cellcolor[HTML]{2080D3}\textbf{100.0}& \cellcolor[HTML]{2080D3}\textbf{100.0} \\
\hline 

\end{tabular}
}
\caption{\label{tab:ft_XLM-RoBERTa_bn}
The table represents the $G_S$ and $S_S$ using \textbf{FT} over fine-tuned \texttt{XLM-RoBERTa} on the fever `\textbf{bn}' dataset. 
}
\end{table*} 
%%%%%%%%%%%%%%%%%%%%%%%% FT-XLM-RoBERTa-mixed %%%%%%%%%%%%%%%%%%%%%%%%%%%% 
\begin{table*}[htpb]
\centering

\resizebox{\linewidth}{!}{%
\begin{tabular}{>{\hspace{0pt}}m{0.020\linewidth}>{\hspace{0pt}}m{0.045\linewidth}|>{\centering\hspace{0pt}}m{0.069\linewidth}>{\centering\hspace{0pt}}m{0.069\linewidth}>{\centering\hspace{0pt}}m{0.069\linewidth}>{\centering\hspace{0pt}}m{0.069\linewidth}>{\centering\hspace{0pt}}m{0.069\linewidth}>{\centering\hspace{0pt}}m{0.069\linewidth}|>{\centering\hspace{0pt}}m{0.058\linewidth}>{\centering\hspace{0pt}}m{0.058\linewidth}>{\centering\hspace{0pt}}m{0.058\linewidth}>{\centering\hspace{0pt}}m{0.069\linewidth}>{\centering\hspace{0pt}}m{0.069\linewidth}>{\centering\arraybackslash\hspace{0pt}}m{0.069\linewidth}} 
\hline
 &  & \multicolumn{6}{>{\centering\hspace{0pt}}m{0.39\linewidth}|}{$G_S$  ($x'$) $\rightarrow$} & \multicolumn{6}{>{\centering\arraybackslash\hspace{0pt}}m{0.39\linewidth}}{$S_S$ ($\hat{x}$) $\rightarrow$} \\ 
\hline\multicolumn{1}{>{\centering\hspace{0pt}}m{0.05\linewidth}}{\textbf{Set}} & \textbf{$x$} $\downarrow$ & \textbf{en} & \textbf{fr} & \textbf{es} & \textbf{hi} & \textbf{gu} & \textbf{bn} & \textbf{en} & \textbf{fr} & \textbf{es} & \textbf{hi} & \textbf{gu} & \textbf{bn} \\ \hline\multicolumn{1}{>{\centering\hspace{0pt}}m{0.05\linewidth}}{\textbf{IL}}& en & \cellcolor[HTML]{3189D6}\textbf{95.39}& \cellcolor[HTML]{67A8E1}80.64& \cellcolor[HTML]{69A9E1}80.22& \cellcolor[HTML]{D6E7F6}50.96& \cellcolor[HTML]{DBEAF8}49.45& \cellcolor[HTML]{DEECF8}48.79& \cellcolor[HTML]{2080D3}\textbf{100.0}& \cellcolor[HTML]{2080D3}\textbf{100.0}& \cellcolor[HTML]{2080D3}\textbf{100.0}& \cellcolor[HTML]{2080D3}\textbf{100.0}& \cellcolor[HTML]{2080D3}\textbf{100.0}& \cellcolor[HTML]{2080D3}\textbf{100.0} \\
\hline 

\end{tabular}
}
\caption{\label{tab:ft_XLM-RoBERTa_mixed}
The table represents the $G_S$ and $S_S$ using \textbf{FT} over fine-tuned \texttt{XLM-RoBERTa} on the fever `\textbf{mixed}' dataset. 
}
\end{table*} 
%%%%%%%%%%%%%%%%%%%%%%%% FT-XLM-RoBERTa-inverse %%%%%%%%%%%%%%%%%%%%%%%%%%%% 
\begin{table*}[htpb]
\centering

\resizebox{\linewidth}{!}{%
\begin{tabular}{>{\hspace{0pt}}m{0.020\linewidth}>{\hspace{0pt}}m{0.045\linewidth}|>{\centering\hspace{0pt}}m{0.069\linewidth}>{\centering\hspace{0pt}}m{0.069\linewidth}>{\centering\hspace{0pt}}m{0.069\linewidth}>{\centering\hspace{0pt}}m{0.069\linewidth}>{\centering\hspace{0pt}}m{0.069\linewidth}>{\centering\hspace{0pt}}m{0.069\linewidth}|>{\centering\hspace{0pt}}m{0.058\linewidth}>{\centering\hspace{0pt}}m{0.058\linewidth}>{\centering\hspace{0pt}}m{0.058\linewidth}>{\centering\hspace{0pt}}m{0.069\linewidth}>{\centering\hspace{0pt}}m{0.069\linewidth}>{\centering\arraybackslash\hspace{0pt}}m{0.069\linewidth}} 
\hline
 &  & \multicolumn{6}{>{\centering\hspace{0pt}}m{0.39\linewidth}|}{$G_S$  ($x'$) $\rightarrow$} & \multicolumn{6}{>{\centering\arraybackslash\hspace{0pt}}m{0.39\linewidth}}{$S_S$ ($\hat{x}$) $\rightarrow$} \\ 
\hline\multicolumn{1}{>{\centering\hspace{0pt}}m{0.05\linewidth}}{\textbf{Set}} & \textbf{$x$} $\downarrow$ & \textbf{en} & \textbf{fr} & \textbf{es} & \textbf{hi} & \textbf{gu} & \textbf{bn} & \textbf{en} & \textbf{fr} & \textbf{es} & \textbf{hi} & \textbf{gu} & \textbf{bn} \\ \hline\multicolumn{1}{>{\centering\hspace{0pt}}m{0.05\linewidth}}{\textbf{IL}}& en & \cellcolor[HTML]{CCE2F4}\textbf{53.65}& \cellcolor[HTML]{D2E5F6}52.05& \cellcolor[HTML]{D2E5F6}52.05& \cellcolor[HTML]{D2E5F6}52.05& \cellcolor[HTML]{D2E5F6}52.05& \cellcolor[HTML]{D2E5F6}52.05& \cellcolor[HTML]{2080D3}\textbf{100.0}& \cellcolor[HTML]{2080D3}\textbf{100.0}& \cellcolor[HTML]{2080D3}\textbf{100.0}& \cellcolor[HTML]{2080D3}\textbf{100.0}& \cellcolor[HTML]{2080D3}\textbf{100.0}& \cellcolor[HTML]{2080D3}\textbf{100.0} \\
\hline 

\end{tabular}
}
\caption{\label{tab:ft_XLM-RoBERTa_inverse}
The table represents the $G_S$ and $S_S$ using \textbf{FT} over fine-tuned \texttt{XLM-RoBERTa} on the fever `\textbf{inverse}' dataset. 
}
\end{table*}

% \begin{table}
% \centering
% \begin{tabular}{lc}
% \hline
% \textbf{Command} & \textbf{Output}\\
% \hline
% \verb|{\"a}| & {\"a} \\
% \verb|{\^e}| & {\^e} \\
% \verb|{\`i}| & {\`i} \\ 
% \verb|{\.I}| & {\.I} \\ 
% \verb|{\o}| & {\o} \\
% \verb|{\'u}| & {\'u}  \\ 
% \verb|{\aa}| & {\aa}  \\\hline
% \end{tabular}
% \begin{tabular}{lc}
% \hline
% \textbf{Command} & \textbf{Output}\\
% \hline
% \verb|{\c c}| & {\c c} \\ 
% \verb|{\u g}| & {\u g} \\ 
% \verb|{\l}| & {\l} \\ 
% \verb|{\~n}| & {\~n} \\ 
% \verb|{\H o}| & {\H o} \\ 
% \verb|{\v r}| & {\v r} \\ 
% \verb|{\ss}| & {\ss} \\
% \hline
% \end{tabular}
% \caption{Example commands for accented characters, to be used in, \emph{e.g.}, Bib\TeX{} entries.}
% \label{tab:accents}
% \end{table}

% \begin{table*}
% \centering
% \begin{tabular}{lll}
% \hline
% \textbf{Output} & \textbf{natbib command} & \textbf{Old ACL-style command}\\
% \hline
% \citep{Gusfield:97} & \verb|\citep| & \verb|\cite| \\
% \citealp{Gusfield:97} & \verb|\citealp| & no equivalent \\
% \citet{Gusfield:97} & \verb|\citet| & \verb|\newcite| \\
% \citeyearpar{Gusfield:97} & \verb|\citeyearpar| & \verb|\shortcite| \\
% \hline
% \end{tabular}
% \caption{\label{citation-guide}
% Citation commands supported by the style file.
% The style is based on the natbib package and supports all natbib citation commands.
% It also supports commands defined in previous ACL style files for compatibility.
% }
% \end{table*}
\end{document}